\documentclass{article} %
\usepackage[preprint]{colm2026_conference}

\usepackage{microtype}
\usepackage{hyperref}
\usepackage{url}
\usepackage{booktabs}
\usepackage{multirow}
\usepackage{graphicx}
\usepackage{enumitem}
\usepackage{fancyvrb}
\usepackage{capt-of}
\usepackage{placeins}

\usepackage{amsmath,amsfonts,bm}

\def\eqref#1{equation~\ref{#1}}

\def\1{\bm{1}}

\DeclareMathAlphabet{\mathsfit}{\encodingdefault}{\sfdefault}{m}{sl}
\SetMathAlphabet{\mathsfit}{bold}{\encodingdefault}{\sfdefault}{bx}{n}

\usepackage{lineno}

\definecolor{darkblue}{rgb}{0, 0, 0.5}
\definecolor{injecthl}{RGB}{255,251,204}
\definecolor{recallhl}{RGB}{204,229,255}
\newcommand{\HL}[1]{\colorbox{injecthl}{#1}}
\newcommand{\RHL}[1]{\colorbox{recallhl}{#1}}
\hypersetup{colorlinks=true, citecolor=darkblue, linkcolor=darkblue, urlcolor=darkblue}

\title{RecaLLM: Addressing the Lost-in-Thought Phenomenon with Explicit In-Context Retrieval}

\author{
Kyle Whitecross, Negin Rahimi \\
\texttt{\{kwhitecross, rahimi\}@cs.umass.edu} \\
University of Massachusetts Amherst
}

\newcommand{\projectname}{RecaLLM}

\begin{document}

\ifcolmsubmission
\linenumbers
\fi

\maketitle

\begin{abstract}

    We propose \textbf{\projectname{}}, a set of reasoning language models post-trained to make effective use of long-context information. In-context retrieval, which identifies relevant evidence from context, and reasoning are deeply intertwined: retrieval supports reasoning, while reasoning often determines what must be retrieved. However, their interaction remains largely underexplored. 
In preliminary experiments on several open-source LLMs, we observe that in-context retrieval performance substantially degrades even after a short reasoning span, revealing a key bottleneck for test-time scaling that we refer to as \textbf{lost-in-thought}:
reasoning steps that improve performance also make subsequent in-context retrieval more challenging.
To address this limitation, \projectname{} \textit{interleaves} reasoning with explicit in-context retrieval, alternating between reasoning and retrieving context information needed to solve intermediate subproblems.
We introduce a negligible-overhead constrained decoding mechanism that enables verbatim copying of evidence spans, improving the grounding of subsequent generation.
Trained on diverse lexical and semantic retrieval tasks, RecaLLM achieves strong performance on two long-context benchmarks, RULER and HELMET, significantly outperforming baselines.
Notably, we observe consistent gains at context windows of up to 128K tokens using training samples of at most 10K tokens, far shorter than those used by existing long-context approaches, highlighting a promising path toward improving long-context performance without expensive long-context training data.\footnote{Code, data, and models available at \url{https://github.com/kswhitecross/RecaLLM}.}

\end{abstract}

\section{Introduction}
\label{sec:intro}

\begin{figure*}[t]
    \centering
    \includegraphics[width=0.9\textwidth]{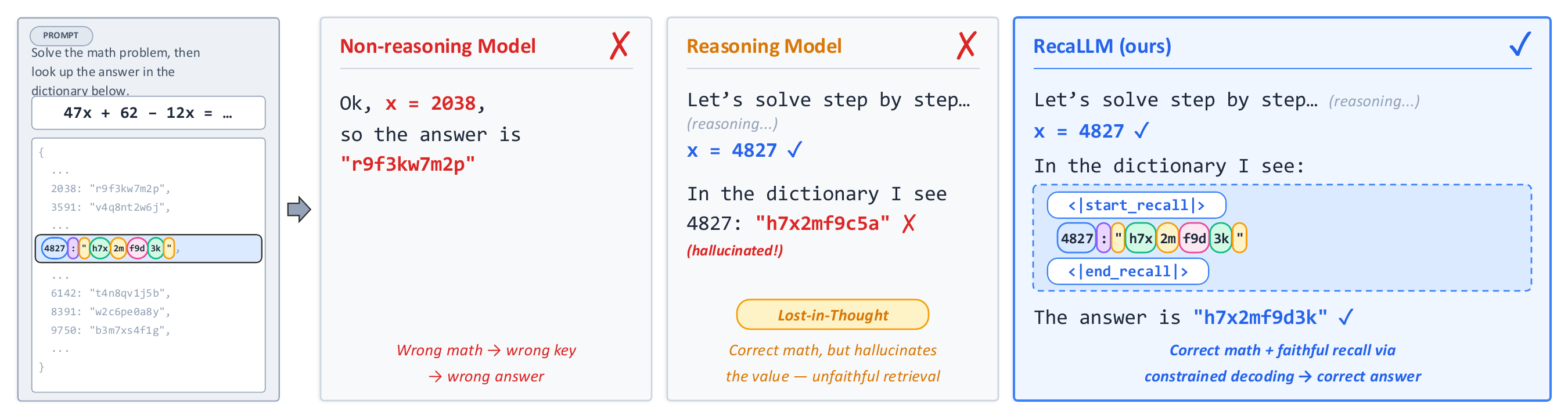}
    \vspace{-0.3cm}
    \caption{
    Illustration of lost-in-thought and how \projectname{} mitigates it with explicit, faithful \emph{recall spans}.
    A reasoning model may recover the correct key yet still hallucinate the value.
    }
    \label{fig:hero}
\end{figure*}

Long-context large language models (LLMs) enable long in-context learning~\citep{bertsch-etal-2025-context}, complex reasoning through scaling test-time compute~\citep{guo2025deepseek,olmo2025olmo}, and long-horizon agentic workflows~\citep{sun-etal-2025-simpledeepsearcher,li2025webthinker,zheng-etal-2025-deepresearcher,openai2025deepresearch,google2025geminideepresearch,yen2025lost}, in addition to supporting a wide range of complex real-world applications. 
However, extending the context window~\citep{lu2025a} does not by itself ensure effective use of long-context information~\citep{hsieh2024ruler,yen2024helmet}.
Prior work shows that LLMs exhibit  positional biases~\citep{liu-etal-2024-lost,li2025longcontext,tian-etal-2025-distance,wu-etal-2025-pandoras} and that their ability to use relevant information degrades as the context length or the difficulty of irrelevant information increases~\citep{pmlr-v202-shi23a,yang-etal-2025-llm-reasoning,wu2024how}. 
Effective long-context use therefore depends critically on \emph{in-context retrieval}, the ability to retrieve relevant information from the input context~\citep{hsieh2024ruler,qiu-etal-2025-eliciting}.
We show that this capability degrades even further in reasoning language models, where long chain-of-thought reasoning, a key driver of their performance, substantially increases the context length with semantically related, but potentially hard distracting tokens. We refer to this failure as \emph{lost-in-thought} (Figure~\ref{fig:hero}).

Retrieval-based approaches to long-context modeling remain limited in scope. Prior work, including LoongRL~\citep{wang2026loongrl} and ALR\textsuperscript{2}~\citep{li2024alr2}, largely trains models to treat in-context retrieval as a step that can be performed before reasoning begins: the initial query is available in the context and is assumed sufficiently clear to retrieve all evidence needed for the task. However, this assumption is often too restrictive for open-ended tasks, where the need for (additional) context information may emerge after several intermediate reasoning steps and therefore cannot be completely planned upfront. To our knowledge, prior work has not explicitly studied retrieval needs that arise dynamically during the reasoning process itself.
This gap is consequential, because effective in-context retrieval in such settings may need to operate not only over the original input context, but also over the model’s previously generated intermediate outputs\textemdash an increasingly important yet still underexplored setting.

We train \projectname{} on a diverse set of tasks ranging from instances that require no retrieval to those that require multiple retrieval steps. For retrieval-intensive cases, we vary context length, the location of relevant evidence, the type of retrieval required, including both lexical and semantic retrieval, and distractor difficulty through random, hard lexical, and hard semantic negatives. We then post-train LLMs in two stages: a supervised cold start on teacher-annotated rollouts, followed by GRPO~\citep{shao2024grpo}. Crucially, explicit recall spans, together with constrained decoding, make retrieval directly verifiable. As a result, \projectname{} rewards not only final answer quality but also successful retrieval of known relevant evidence, going beyond prior approaches~\citep{wang2026loongrl} that mainly optimize outcome reward alone. Ablations show the importance of this fine-grained training signal. 

Across both synthetic and realistic long-context benchmarks, \projectname{} substantially improves effective context use. On RULER~\citep{hsieh2024ruler}, \projectname{}-Qwen2.5-7B achieves the best average among the 7--8B models at 92.8 and remains strong even at 128K tokens, outperforming larger long-context baselines such as LoongRL-14B~\citep{wang2026loongrl}. Relative to the base models, \projectname{} gains generally become larger as context length increases: for Qwen, the improvement grows from +6.8 at 4K to +16.1 at 128K, and for Llama the largest gain also appears at 128K (+24.2). On HELMET~\citep{yen2024helmet}, \projectname{} improves its base models by 16.1--17.7 points on average and attains the strongest overall results in the 7--8B class, especially on retrieval-intensive categories such as Recall, ICL, Re-rank, and citation. Notably, these improvements are not limited to simple key-value lookup: \projectname{}-Llama rises from 3.0 to 64.1 on ICL and from 21.3 to 53.2 on Re-rank, indicating that explicit recall improves not only evidence retrieval but also reasoning over retrieved evidence.

We further show that strong long-context gains can be achieved without training on comparably long sequences. \projectname{} is trained on contexts of at most 10K tokens, yet it yields consistent improvements up to 128K tokens, the longest evaluation length. This is particularly notable given that recent long-context methods often rely on larger training contexts: ProLong~\citep{gao2025prolong} is trained on sequences up to 512K tokens, exceeding its 128K evaluation length, while LoongRL~\citep{wang2026loongrl} trains at 16K. These results suggest that directly improving in-context retrieval can yield long-context performance that generalizes well beyond the training range. Given the substantial cost of long-context training and the difficulty of curating high-quality long-sequence data, our approach points to a more efficient path for extending the effective context length of LLMs.

We summarize our main contributions as follows: (1)~We identify \emph{lost-in-thought}, a failure mode in reasoning LLMs where long chain-of-thought makes in-context retrieval harder. (2)~We propose \projectname{}, which interleaves reasoning with explicit in-context retrieval through \emph{recall spans}. (3)~We introduce context-aware constrained decoding for recall spans, ensuring faithful verbatim recall from context and making retrieval directly verifiable. (4)~We show that \projectname{} delivers strong long-context gains, especially on retrieval-intensive tasks, and generalizes from 10K training contexts to evaluations up to 128K.

\section{Lost-in-Thought: How Reasoning Degrades In-Context Retrieval}
\label{sec:problem}

To study how reasoning interacts with long-context capabilities in modern LLMs, we construct a synthetic benchmark that isolates in-context retrieval performance before and after reasoning.  The benchmark centers on a large structured key-value dictionary provided as part of the prompt, and defines two tasks:

\begin{itemize}[noitemsep,topsep=0pt,leftmargin=*]
    \item \textbf{Retrieval}: the prompt directly specifies a query key and asks the model to retrieve the corresponding value, following prior synthetic retrieval benchmarks such as RULER~\citep{hsieh2024ruler}.
    \item \textbf{Reasoning-Retrieval}: the prompt instead presents a math problem whose solution determines the query key, requiring the model to first reason and then retrieve.
\end{itemize}

Context lengths range from 4K to 128K tokens, controlled by varying the number of distractor key-value pairs.  To improve robustness, we vary prompt templates, query placement relative to the dictionary, math problem type, and dictionary format (CSV, JSON, list); all other factors are held fixed across length conditions.  Examples are in Appendix~\ref{app:section_2_examples}.

\begin{figure*}[t]
    \centering
    \includegraphics[width=0.9\textwidth]{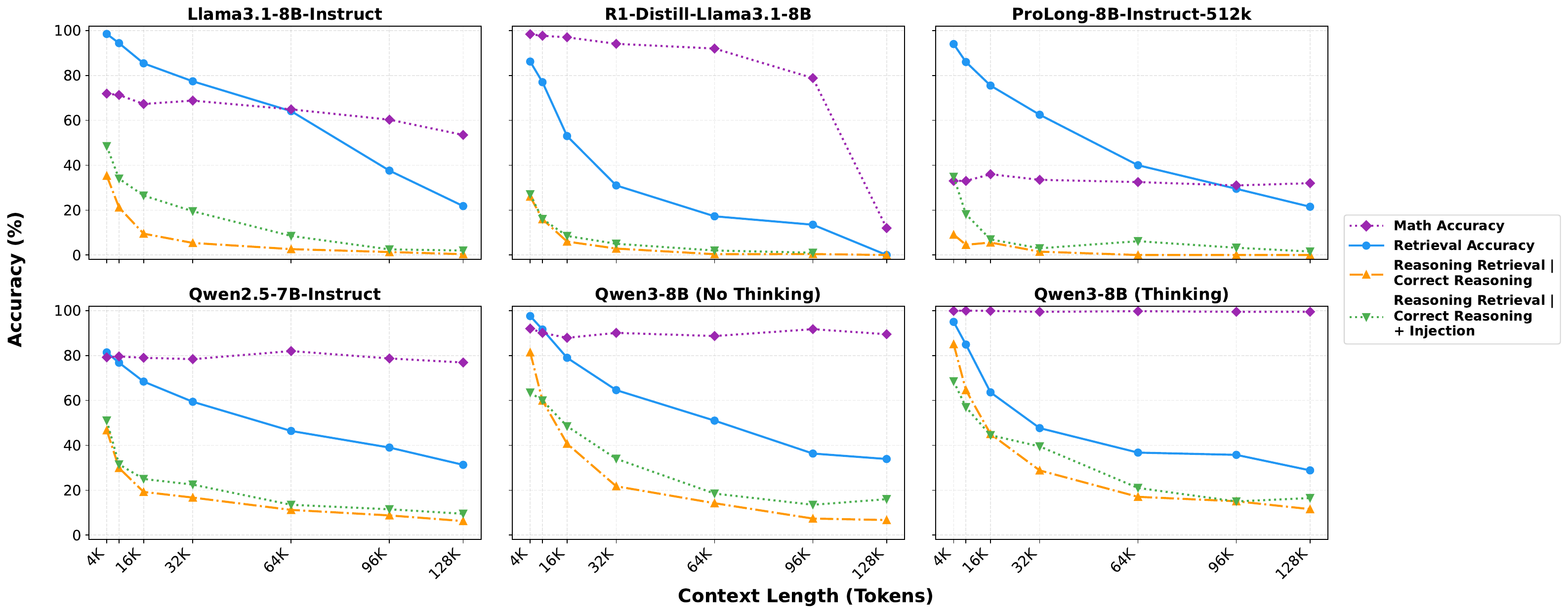}
    \vspace{-0.3cm}
    \caption{
    \textit{Lost in thought}: retrieval accuracy before and after reasoning.
    Injected accuracy measures faithful copying after providing the correct key and prefix.
    }
    \label{fig:retrieval-gap}
\end{figure*}

Figure~\ref{fig:retrieval-gap} presents results for five open-source 7--8B LLMs: \textbf{Llama-3.1-8B-Instruct}~\citep{dubey2024llama3}, \textbf{R1-Distill-Llama-8B}~\citep{guo2025deepseek}, \textbf{ProLong-8B-512K}~\citep{gao2025prolong}, \textbf{Qwen2.5-7B-Instruct}~\citep{qwen2025qwen25}, and \textbf{Qwen3-8B}~\citep{yang2025qwen3} (details in Appendix~\ref{app:section_2_llms}). Across all models, retrieval accuracy after reasoning is substantially worse than direct retrieval, even though math accuracy remains high and stable.  This gap persists across all context lengths (Table~\ref{tab:prompt-injection}).  We refer to this phenomenon as \textit{lost in thought}.

A follow-up injection experiment (Appendix~\ref{app:section_2_injection}) confirms that this degradation stems largely from the model’s inability to faithfully \emph{copy} information from context after reasoning, rather than from failing to identify what to retrieve: even when given the correct key and its exact lexical prefix mid-generation, models still frequently hallucinate the value (Figure~\ref{fig:retrieval-gap}, Table~\ref{tab:prompt-injection}).  We hypothesize that the semantically related tokens generated during reasoning act as distractors that interfere with faithful reproduction.

\section{\projectname}
\label{sec:method}

To address lost-in-thought, \projectname{} makes in-context retrieval an explicit step within generation.  Models reason until they identify a need for contextual evidence, then enter a \emph{recall span} that copies that evidence verbatim from the input or prior generation, turning a long-context retrieval problem into a local reasoning one.  Two components enable this: special delimiter tokens that mark recall spans, and a constrained decoding mechanism that restricts generation within them to valid continuations of the searchable context, guaranteeing faithfulness by construction.  The model decides \emph{what} and \emph{when} to recall; constrained decoding ensures the recalled content is exact; and RL training (Section~\ref{sec:experimental-setup}) teaches the model to use this capability effectively.

\subsection{Recall Tokens and Recall Spans}
\label{sec:recall-tokens}

We extend the base vocabulary $\mathcal{V}$ with two special tokens: \texttt{<|start\_recall|>} ($R_{\text{start}}$) and \texttt{<|end\_recall|>} ($R_{\text{end}}$), which denote the beginning and end of a recall span, respectively.  Let $x = (x_1, \ldots, x_n)$ denote the tokenized input prompt.  The model generates an output sequence over the augmented vocabulary $\mathcal{V} \cup \{R_{\text{start}}, R_{\text{end}}\}$, consisting of regular output tokens interleaved with recall spans:
\[
(\underbrace{y_1, \ldots, y_T}_{\text{reasoning}},\; R_{\text{start}},\; \underbrace{r_1, \ldots, r_K}_{\text{recall span}},\; R_{\text{end}},\; \underbrace{y_{T+1}, \ldots}_{\text{reasoning}},\; \ldots\,)
\]
where $y_1, \ldots, y_T$ are regular output tokens generated before the recall span and $r_1, \ldots, r_K$ are tokens generated within the recall span.  Multiple recall spans may appear in a single generation.

Outside of recall spans, generation proceeds under the standard next-token distribution: $y_t \sim p_\theta(\,\cdot \mid x, y_{<t})$.
When the model emits $R_{\text{start}}$, it enters a recall span.  We define the \emph{searchable context} $c$ as the concatenation of the input prompt with all tokens generated prior to the current recall span: $c = x \mathbin{\|} y_{\leq T}$.
Let $r_{1:k}$ denote the $k$ tokens generated so far inside the current recall span. The \emph{valid continuation set} is
\[
\mathcal{A}(c, r_{1:k}) =
\left\{
v \in \mathcal{V}
\;\middle|\;
\exists\, i \text{ such that } c_{i:i+k-1} = r_{1:k} \text{ and } c_{i+k} = v
\right\},
\]
the set of all tokens that could continue at least one exact occurrence of the recalled prefix in the searchable context.  During a recall span, the model may only emit tokens from this set, or the stop token $R_{\text{end}}$.  Formally, if $z_v$ denotes the model's logit for token $v$, we apply the mask
\[
\tilde{z}_v =
\begin{cases}
z_v & \text{if } v \in \mathcal{A}(c, r_{1:k}) \cup \{R_{\text{end}}\}, \\
-\infty & \text{otherwise},
\end{cases}
\]
and sample the next token from the softmax over $\tilde{z}$.  This guarantees that \emph{every recall span is a contiguous substring of the searchable context}.  
This constrained decoding does not guarantee relevance, but by faithfully copying evidence into the generation, it enables the model to reason over grounded evidence in subsequent steps.
Learning \emph{which} spans to recall, and \emph{when}, is the role of RL training (Section~\ref{sec:experimental-setup}). Because the logit mask does not modify the model's internal representations and is differentiable with respect to the allowed logits, RL training proceeds similarly to the unconstrained case.
The logit mask depends only on token IDs and is computed on a separate CPU thread, fully overlapped with the GPU forward pass, adding negligible overhead to generation latency; implementation details and complexity analysis are provided in Appendix~\ref{app:efficient_implementation}.

\section{\projectname{} Training}
\label{sec:experimental-setup}

We train \projectname{} using a two-stage pipeline similar to~\citet{guo2025deepseek}: a Supervised Finetuning~(SFT) cold start on a small set of reasoning traces annotated with recall tokens, followed by RL on a diverse task mixture.

\subsection{Supervised Finetuning Cold Start}

We train the model with a two-stage adaptation procedure that first learns the new token embeddings and subsequently performs brief full-model finetuning, encouraging the model to use recall tokens naturally during interleaved generation (Appendix~\ref{app:sft_training_details}).

To construct the SFT dataset, we collect reasoning traces from six teacher models across four reasoning and retrieval tasks. Each teacher is prompted to reason and explicitly reference relevant context information, and we retain only completions that produce the correct answer. We then prompt GPT-5.2~\citep{openai2025gpt5} to rewrite these traces so that references to context are realized as verbatim recall spans, providing the annotator with gold documents to ensure recall span accuracy. After annotation, we align each recall span to its source text via fuzzy string matching and discard failed alignments, yielding 1,795 properly annotated examples (Appendix~\ref{app:sft_annotation}).

\newcommand{\cell}[1]{\begin{tabular}[t]{@{}l@{}}#1\end{tabular}}

\begin{table}[t]
\centering
\small
\setlength{\tabcolsep}{4pt}
\begin{tabular}{@{}llrc@{}}
\toprule
\textbf{Category} & \textbf{Datasets} & \textbf{Count} & \textbf{Context Length} \\
\midrule
Multi-Hop QA          & HotpotQA, MuSiQue, 2WikiMQA  & 3,000 & 8--10K    \\
Single-Hop QA         & NQ, TriviaQA                  & 2,000 & 8--10K    \\
Retrieval             & KV Retrieval, Multi-NIAH      & 2,000 & 8--10K    \\
Reasoning Retrieval   & Math Retrieval                & 2,000 & 8--10K    \\
Short-Context Math    & DAPO Math, MCQA Math          & 1,000 & $\leq$1K  \\
In-Context Learning   & Banking77, MASSIVE            & 3,000 & 8--10K    \\
Long-Document QA      & QuALITY                       & 1,000 & $\leq$10K \\
Aggregation       & Majority Vote, Top-N Vote     & 2,000 & 8--10K    \\
Reranking         & MSMARCO v2                    & 2,000 & 8--10K    \\
Entity Citation   & QAMPARI                       & 2,000 & 8--10K    \\
\bottomrule
\end{tabular}
\caption{RL training dataset composition. Detailed descriptions in Appendix~\ref{app:rl_dataset_details}.}
\label{tab:dataset-mix}
\end{table}

\subsection{RL Training Data}

We further train the SFT checkpoints with GRPO~\citep{shao2024grpo} on a broad but intentionally shallow mixture of 20,000 examples across 10 task categories (Table~\ref{tab:dataset-mix}), encouraging the model to learn strategic in-context recall rather than task-specific heuristics.
The categories vary along several axes of recall difficulty. Single-hop and multi-hop QA present identical-looking contexts but require different amounts of retrieval, requiring the model to interleave reasoning and retrieval strategically rather than following a fixed pattern. Passage reranking requires selective recall under a limited generation budget. Short-context math and two novel synthetic aggregation tasks (majority vote and top-$N$ vote) require no retrieval, teaching the model to invoke recall only when beneficial, following~\citet{wang2026loongrl}.
To prevent \projectname{} from learning recall behavior tied to narrow surface patterns, we apply systematic data augmentation across all categories, varying instruction phrasing, question placement, passage format (e.g., fixed-window chunks vs.\ natural paragraphs, with or without titles), and distractor composition, sampling from random, BM25~\citep{robertson1995okapi}, and dense retrieval~\citep{zhang2024mgte} negatives.
Dataset descriptions and augmentation details are provided in Appendix~\ref{app:rl_dataset_details}.

\subsection{RL Reward Function}
\label{sec:reward-function}

A central challenge in our RL training is that successful behavior is inherently compositional: the model must first accurately retrieve supporting evidence through recall spans and then use that evidence to generate the correct final answer. We therefore optimize a composite reward that jointly scores formatting, answer quality, and evidence recall quality:
\begin{align}
R &= 0.2 \cdot R_{\text{format}} + 0.4 \cdot R_{\text{add}} + 0.4 \cdot R_{\text{mult}},
\label{eq:reward} \\
R_{\text{add}} &= 0.5 \cdot R_{\text{ans}} + 0.5 \cdot R_{\text{ret}}, \\
R_{\text{mult}} &= \sqrt{(R_{\text{ans}} + \epsilon)(R_{\text{ret}} + \epsilon)} - \epsilon \qquad (\epsilon = 0.01),
\end{align}
where $R_{\text{format}}$ verifies the required interleaved output format, $R_{\text{ans}}$ is a task-specific answer quality metric such as exact match, F1, or NDCG@10 (Table~\ref{tab:rl-config}), and $R_{\text{ret}}$ measures recall quality against gold evidence.
The additive component $R_{\text{add}}$ provides a learning signal when either the final answer or the recalled evidence is (partially) correct, which is especially important early in training when jointly successful trajectories are sparse.
In contrast to purely additive reward formulations~\citep{wang2026loongrl,jin2025searchr1,song2025r1searcher}, we introduce $R_{\text{mult}}$, a smoothed geometric mean that explicitly favors rollouts achieving high answer quality and high retrieval quality simultaneously.

\paragraph{In-context Retrieval Reward.}
For tasks with gold passages, $R_{\text{ret}}$ measures how well the model's recall spans recover the gold evidence. 
Because both gold passages and recalled spans generated under constrained decoding are contiguous substrings of the context, each can be identified by its character-level position interval.
We compute an F1 score over the intersection of these intervals (Appendix~\ref{app:retrieval_reward_formal}), normalized by a task-specific hit threshold $\tau$ that caps the reward once overlap is sufficient. This prevents the reward from favoring exhaustive copying and reflects the fact that for many tasks (e.g., multi-hop or single-hop QA) only a small portion of the gold passage is actually relevant to the question.
These per-passage scores are then averaged: over all gold passages for tasks with few gold documents (e.g., multi-hop QA), or over the top-$K$ highest-scoring passages for tasks with many (e.g., reranking).
Importantly, this interval-based reward is only precise because constrained decoding guarantees verbatim reproduction; without it, even a single non-verbatim character breaks the contiguous match, making the signal substantially noisier.
For tasks without segmented gold evidence (long-document QA), the overlap score is set to 1 if any recall span is present; for tasks that do not require retrieval (short-context math, aggregation), it is set to 1 unconditionally.
Formal definitions and per-category reward configurations are in Appendix~\ref{app:retrieval_reward_formal} and Table~\ref{tab:rl-config}.

\paragraph{Penalties.}
To prevent pathological recall behavior, $R_{\text{ret}}$ is further modulated by two penalties. A \emph{density penalty} exponentially downweights reward when the frequency of recall spans exceeds a threshold, with a task-dependent number of initial spans exempt (Table~\ref{tab:rl-config}). A \emph{correctness penalty} detects malformed recall spans, such as very short spans or mismatched start/end delimiter tokens. Formal definitions are in Appendix~\ref{app:retrieval_reward_formal}.
We optimize this reward using GRPO~\citep{shao2024grpo} on 4 A100 GPUs for a single epoch with 16 rollouts per example. Training infrastructure and hyperparameters are detailed in Appendix~\ref{app:rl_training_details}.

\section{Experimental Setup and Results}
\label{sec:results}

We evaluate \projectname{} across two base model families to test the robustness of its gains across architectures. Specifically, we train \projectname{} variants initialized from Qwen2.5-7B-Instruct~\citep{qwen2025qwen25} and Llama-3.1-8B-Instruct~\citep{dubey2024llama3}. Appendix~\ref{sec:training_dynamics} provides training-curve analyses showing stable optimization, increasingly selective and accurate recall, and consistent gains across all ten tasks in Table~\ref{tab:dataset-mix} for both \projectname{} models.

In addition to evaluating \projectname{} models on validation splits of our training datasets, we assess their generalizability  on two established long-context benchmarks: RULER~\citep{hsieh2024ruler} and HELMET~\citep{yen2024helmet}, following prior work on long-context evaluation~\citep{gao2025prolong,wang2026loongrl,olmo2025olmo}. More details of these two benchmarks are provided in Appendix~\ref{app:eval_benchmarks}.

\begin{figure*}[t]
    \centering
    \includegraphics[width=\textwidth]{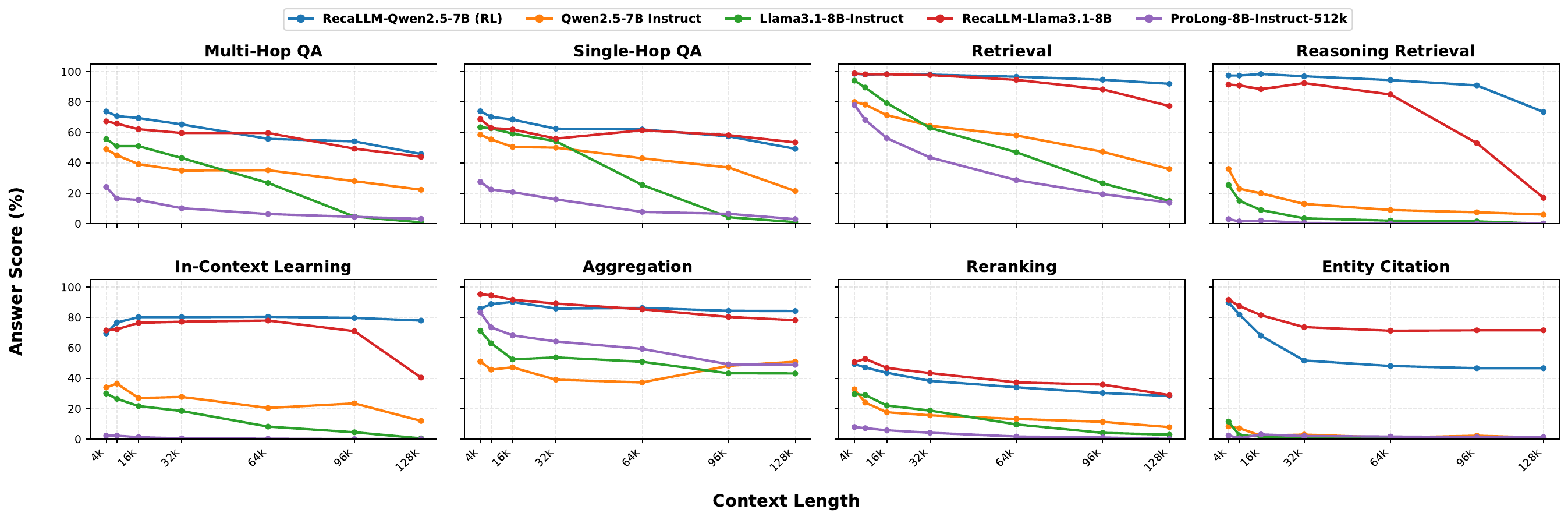}
    \vspace{-0.8cm}
    \caption{Per-category answer scores across context lengths (4K--128K) on the in-domain evaluation. \projectname{} models maintain strong performance as context length increases, while baselines degrade sharply beyond 32K tokens.}
    \label{fig:indomain-scaling}
\end{figure*}

\subsection{Performance on In-Domain Long-Context Tasks}

Figure~\ref{fig:indomain-scaling} and Table~\ref{tab:indomain} in Appendix~\ref{app:indomain_table} report performance on validation splits of the training datasets across context lengths from 4K to 128K.  
\projectname{} models substantially outperform all baselines across nearly every category, with the largest gains emerging at longer contexts, where baseline performance degrades sharply.  
The improvements are particularly significant for tasks that require retrieval after reasoning. 
On reasoning-retrieval, \projectname{}-Qwen improves from 23.0\% to 97.6\% at short contexts and from 7.5\% to 86.3\% at long contexts. On entity citation, \projectname{}-Llama improves from 4.1\% to 83.6\% at short contexts and from 0.3\% to 71.5\% at long contexts.
Consistent gains also appear on reasoning-oriented tasks: short-context math improves by 5.2 points for \projectname{}-Qwen, and aggregation benefits substantially despite not requiring explicit recall.

\subsection{Performance on Long-Context Benchmarks}

\begin{table*}[t]
\centering
\footnotesize
\setlength{\tabcolsep}{3.5pt}
\resizebox{\textwidth}{!}{%
\begin{tabular}{@{}l rrrrrr r rrrrrrr r@{}}
\toprule
& \multicolumn{7}{c}{\textbf{RULER}} & \multicolumn{8}{c}{\textbf{HELMET}} \\
\cmidrule(lr){2-8} \cmidrule(lr){9-16}
\textbf{Model} & \textbf{4K} & \textbf{8K} & \textbf{16K} & \textbf{32K} & \textbf{64K} & \textbf{128K} & \textbf{Avg} & \textbf{Recall} & \textbf{RAG} & \textbf{ICL} & \textbf{Cite} & \textbf{Re-rank} & \textbf{LongQA} & \textbf{Summ} & \textbf{Avg} \\
\midrule
ProLong-8B-Instruct-512K & 92.3 & 86.0 & 73.7 & 59.5 & 60.5 & 46.8 & 69.8 & 74.3 & 68.7 & 18.6 & 12.4 & 12.5 & \textbf{36.4} & 36.5 & 37.0 \\
QwenLong-L1-32B             & 91.7 & 88.5 & 87.6 & 86.8 & 80.6 & 70.2 & 84.2 & ---  & 52.9 & ---  & 14.3 & ---  & ---  & 47.0 & ---  \\
LoongRL-7B                  & 95.1 & 94.3 & 93.4 & 91.4 & 86.2 & 76.8 & 89.5 & ---  & 61.6 & ---  & 11.4 & ---  & ---  & \textbf{44.5} & ---  \\
LoongRL-14B                 & 97.6 & 96.1 & 95.4 & 95.1 & 87.1 & 79.9 & 91.9 & ---  & 63.7 & ---  & 18.7 & ---  & ---  & 49.8 & ---  \\
\midrule
Llama-3.1-8B-Instruct      & 95.7 & 93.5 & 89.8 & 85.6 & 78.6 & 42.2 & 80.9 & 77.0 & 63.0 & 3.0  & 10.0 & 21.3 & 36.3 & 39.2 & 35.7 \\
\projectname{}-Llama-3.1-8B & 97.7 & \textbf{97.2} & 95.5 & 90.7 & 78.6 & 66.4 & 87.7 & \textbf{98.2} & 68.1 & 64.1 & \textbf{24.7} & \textbf{53.2} & 34.8 & 31.0 & 53.4 \\
\midrule
Qwen2.5-7B-Instruct        & 92.0 & 86.8 & 86.1 & 82.3 & 76.8 & 66.8 & 81.8 & 58.8 & 58.6 & 41.7 & 10.3 & 17.7 & 33.2 & 41.3 & 37.4 \\
\projectname{}-Qwen2.5-7B   & \textbf{98.8} & 96.7 & \textbf{95.8} & \textbf{93.1} & \textbf{89.7} & \textbf{82.9} & \textbf{92.8} & 96.2 & \textbf{69.4} & \textbf{69.7} & 13.7 & 46.2 & 35.5 & 44.0 & \textbf{53.5} \\
\bottomrule
\end{tabular}%
}
\caption{Performance on RULER~\citep{hsieh2024ruler} (averaged across tasks) and HELMET~\citep{yen2024helmet} (averaged across context lengths). Models with --- indicate results not reported in their papers.}
\label{tab:benchmarks}\label{tab:ruler}\label{tab:helmet}
\end{table*}

We evaluate \projectname{} models on the RULER~\citep{hsieh2024ruler} and HELMET~\citep{yen2024helmet} benchmarks, comparing against their corresponding base LLMs as well as ProLong~\citep{gao2025prolong}, LoongRL~\citep{wang2026loongrl}, and QwenLong~\citep{wan2025qwenlong}. For LoongRL and QwenLong, we use the results reported by \citet{wang2026loongrl} whenever available, since LoongRL’s models and implementation are not publicly released. We reproduce results for base LLMs Llama3.1-8B-Instruct and Qwen2.5-7B-Instruct, as well as ProLong-8B-512k under the same reasoning-oriented system prompt used for \projectname{}. To ensure a fair comparison between reasoning and non-reasoning models, we evaluate all  RULER and HELMET tasks using chat templates, 
rather than relying on answer-only prompts that can disadvantage reasoning models.

\paragraph{RULER Results.}
On RULER (Table~\ref{tab:benchmarks}), both \projectname{} models substantially outperform their base models across all context lengths.  
Two patterns are especially notable in Table~\ref{tab:ruler}. First, \projectname{} yields strong parameter-efficient long-context performance: \projectname{}-Qwen achieves the best average score in the 7B--8B class (\textbf{bolded} in Table~\ref{tab:benchmarks}) and even surpasses substantially larger baselines, LoongRL-14B and QwenLong-L1-32B, despite being trained on substantially shorter contexts (8--10K tokens, compared with 16K for LoongRL and up to 60K for QwenLong-L1). 
Second, the gains become larger as context length increases, indicating that \projectname{} improves not only overall accuracy but also robustness to extreme context length. For example, \projectname{}-Qwen improves over its base by 6.8 points at 4K but by 16.1 at 128K, and reduces the 4K-to-128K degradation from 25.2 points to 15.9. Notably, these improvements do not come at the expense of short-context performance, as both RecaLLM variants also outperform their base models at short contexts such as 4K and 8K. 
The sharper decline of \projectname{}-Llama beyond 64K, compared to \projectname{}-Qwen, suggests that the long-context generalization is partly inherited from pretraining.

\paragraph{HELMET Results.}
On HELMET (Table~\ref{tab:benchmarks}), \projectname{} achieves the highest overall average among 7--8B models, improving over its base Qwen2.5-7B and Llama-3.1-8B models by 16.1 and 17.7 points, respectively. This suggests that the gains from \projectname{} are robust across different base model families.
The improvements are especially significant on Recall, ICL, Re-rank, and Cite, where faithful access to dispersed context information is essential. Notably, \projectname{}-Llama improves from 3.0 to 64.1 on ICL and from 21.3 to 53.2 on Re-rank. HELMET's ICL tasks use randomly generated numeric labels rather than semantic class names, testing pure pattern matching over the provided examples. The recall mechanism is particularly well suited to this setting, as it enables the model to faithfully retrieve exact example--label pairs from context rather than hallucinating plausible numbers. These improvements suggest that explicit recall spans enhance not only in-context evidence retrieval, but also downstream reasoning over retrieved evidence.

At the same time, gains on LongQA and Summ are smaller and less consistent. \projectname{}-Qwen improves slightly on both tasks, whereas \projectname{}-Llama maintains LongQA performance but declines on Summ. Both categories require long-form generation and are evaluated with a LLM-as-a-judge. Taken together with the near-saturated Recall performance in Table~\ref{tab:benchmarks}, these results suggest that the remaining challenge in LongQA and Summ lies less in retrieving relevant evidence than in composing that evidence into grounded, coherent long-form outputs.

\section{Analysis}
\label{sec:analysis}

\subsection{Generalizability of the Recall Capability}
\label{sec:policy-generalization}

\begin{figure}[t]
    \centering
    \includegraphics[width=0.8\columnwidth]{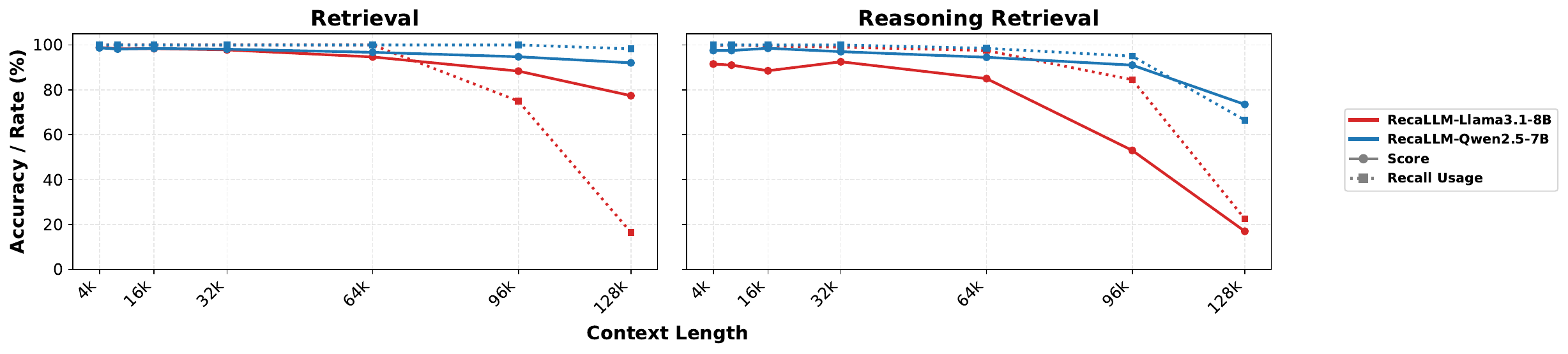}
    \vspace{-0.3cm}
    \caption{Accuracy and recall token usage rate of \projectname{} models across context lengths on the retrieval and reasoning-retrieval tasks from Section~\ref{sec:problem}.}
    \label{fig:retrieval-recall-usage}
\end{figure}

Although \projectname{} is trained on only 8--10K contexts, it generalizes substantially beyond this range. 
Figure~\ref{fig:retrieval-recall-usage} plots task score alongside recall usage rate across context lengths on the retrieval and reasoning-retrieval tasks from Section~\ref{sec:problem}. The results show that accuracy closely tracks recall usage on both tasks: performance remains strong when recall is used consistently, and degrades precisely when recall usage falls. 
\projectname{}-Qwen maintains high recall usage over nearly the entire range and correspondingly sustains strong performance, while \projectname{}-Llama shows a clear drop in recall usage at 96K and 128K, with much larger accuracy drops, especially on reasoning-retrieval.
When recall usage drops, the model reverts to implicit retrieval.
This pattern indicates that a main failure reason at extreme lengths is policy drift away from invoking recall.

\subsection{Ablation Studies}
\label{sec:ablations}

\begin{table}[t]
\centering
\small
\resizebox{\columnwidth}{!}{%
\begin{tabular}{@{}lrrrrrrrrrrr@{}}
\toprule
\textbf{Model} & \textbf{1-Hop} & \textbf{N-Hop} & \textbf{Ret.} & \textbf{R$\to$Ret.} & \textbf{Math} & \textbf{ICL} & \textbf{LQA} & \textbf{Agg.} & \textbf{Rerank} & \textbf{Ent.\ Cite} & \textbf{Avg} \\
\midrule
\projectname{}-Qwen2.5-7B & 63.4 & 62.2 & \textbf{96.7} & \textbf{92.8} & 61.0 & 77.9 & 72.2 & 86.5 & \textbf{38.8} & \textbf{61.8} & \textbf{71.3} \\
\quad -- No Recall Reward      & 53.8 & 54.8 & 94.8 & 88.6 & \textbf{62.0} & 75.8 & \textbf{76.0} & 70.1 & 28.0 & 51.6 & 65.5 \\
\quad -- No Logit Masking           & \textbf{69.5} & \textbf{70.2} & 89.5 & 72.1 & 58.2 & \textbf{78.4} & 72.2 & \textbf{90.6} & 35.3 & 58.1 & 69.4 \\
\bottomrule
\end{tabular}%
}
\caption{Performance of ablated variants of \projectname{}-Qwen on validation sets, averaged over all context lengths.
}
\label{tab:ablations}
\end{table}

To isolate the roles of constrained decoding (Section~\ref{sec:recall-tokens}) and the retrieval reward (Section~\ref{sec:reward-function}), we train two ablations of \projectname{}-Qwen2.5-7B. 
\textbf{No Recall Reward} removes supervision for retrieving the correct evidence, including the density and correctness penalties, and sets $R_{\text{ret}}=1$ unconditionally during RL. 
\textbf{No Logit Masking} ablation keeps the full training recipe, including SFT cold start, GRPO training using the same data and reward model, but disables constrained decoding within recall spans.
Table~\ref{tab:ablations}  reports  results on the validation sets. Additional results on this evaluation as well as  results on RULER and HELMET in Appendix~\ref{app:ablation_benchmarks} show the same overall pattern.

Table~\ref{tab:ablations} indicates that the two components are complementary. Removing the recall reward causes the larger overall degradation, reducing the average score from 71.3 to 65.5, with especially large drops on aggregation, reranking, entity citation, and QA. This suggests that the recall reward is the main training signal that teaches the model to use recall broadly rather than only in tasks where copying is obviously beneficial. 
Removing logit masking leads to a smaller average decline, from 71.3 to 69.4, but causes much larger losses on the retrieval-intensive categories that require exact lexical matching, namely retrieval and reasoning-retrieval.
Meanwhile, removing logit masking improves performance on categories where exact lexical retrieval is less critical, suggesting that unconstrained generation offers additional flexibility on these tasks. 
At the same time, constrained decoding remains important because it makes the recall reward straightforward to incorporate, and that reward has a substantial impact on \projectname{} performance.

\section{Related Work}
\label{sec:related-work}

\paragraph{RL for long-context reasoning.}
LoongRL~\citep{wang2026loongrl} and QwenLong-L1~\citep{wan2025qwenlong} train models via RL to reason over long contexts (Appendix~\ref{app:rl_long_context}), but differ from \projectname{} in three key respects.
First, neither provides an explicit in-context retrieval step nor a mechanism for \emph{faithful} retrieval.
Second, neither explicitly rewards retrieval quality.
Third, both train on substantially longer contexts for far more steps, yet \projectname{}-Qwen2.5-7B outperforms them on RULER and HELMET (Table~\ref{tab:benchmarks}).

\paragraph{Retrieval-augmented reasoning and context extension.}
Agentic RAG models~\citep{jin2025searchr1,song2025r1searcher,li2025webthinker,zheng-etal-2025-deepresearcher} query external corpora during reasoning (Appendix~\ref{app:improving_long_context}). 
\projectname{} is complementary: agentic search systems that retrieve documents into context would benefit from stronger in-context utilization of the evidence they gather.
A large body of work targets context extension at pretraining or continued-training time~\citep{lu2025a,xiong-etal-2024-effective}, including ProLong~\citep{gao2025prolong} and YaRN~\citep{peng2024yarn}. \projectname{} operates at the post-training stage and is agnostic to the underlying context extension recipe.

\paragraph{Constrained decoding and copy mechanisms.}
Grammar-constrained decoding~\citep{willard2023outlines,dong2024xgrammar} and entity-constrained generation~\citep{decao2021genre} use logit masking to enforce structural constraints from a fixed grammar or candidate set (Appendix~\ref{app:constrained_decoding}). 
Classical copy mechanisms~\citep{gu2016copynet,see2017pointer} similarly rely on learned copy distributions.

\section{Conclusion}
\label{sec:conclusion}

We introduce \projectname{}, a family of reasoning language models that address the \emph{lost-in-thought} problem, 
by interleaving reasoning with explicit, constrained-decoding recall spans, achieving strong results on RULER and HELMET that surpass larger models trained on longer contexts in fewer steps. Looking ahead, we believe explicit in-context retrieval opens several promising directions: more flexible retrieval mechanisms that preserve faithfulness without the rigidity of strict constrained decoding, self-recall over the model's own prior generation for long-horizon reasoning consistency, and scaling to larger, more capable models that should better learn when and how to invoke retrieval. More broadly, making retrieval an explicit, verifiable step within reasoning creates new opportunities for reward design, interpretability, and trustworthy grounding in long-context applications.

\section*{Reproducibility Statement}

Code, training data, and trained \projectname{} model weights are publicly available at \url{https://github.com/kswhitecross/RecaLLM}. \projectname{} models are implemented within the Hugging Face Transformers library~\citep{wolf-etal-2020-transformers}, allowing users to download, run and train them with standard inference pipelines. All base models, training datasets, and evaluation benchmarks used in this work are publicly available. Training hyperparameters, reward configurations, and per-category settings are reported in Appendix~\ref{app:section_4}, and evaluation hyperparameters are reported in Appendix~\ref{app:eval_benchmarks}.

\bibliography{colm2026_conference}

@inproceedings{wolf-etal-2020-transformers,
    title = "Transformers: State-of-the-Art Natural Language Processing",
    author = "Wolf, Thomas et al.",
    booktitle = "Proceedings of the 2020 Conference on Empirical Methods in Natural Language Processing: System Demonstrations",
    year = "2020",
    publisher = "Association for Computational Linguistics",
    url = "https://www.aclweb.org/anthology/2020.emnlp-demos.6",
    pages = "38--45"
}

@misc{hewitt2021embeddings,
  author = {Hewitt, John},
  title = {Initializing New Word Embeddings for Pretrained Language Models},
  year = {2021},
  howpublished = {\url{https://www.cs.columbia.edu/~johnhew/vocab-expansion.html}},
  note = {Blog post}
}

@article{olmo2025olmo,
  title={Olmo 3},
  author={Olmo, Team and Ettinger, Allyson and Bertsch, Amanda and Kuehl, Bailey and Graham, David and Heineman, David and Groeneveld, Dirk and Brahman, Faeze and Timbers, Finbarr and Ivison, Hamish and others},
  journal={arXiv preprint arXiv:2512.13961},
  year={2025}
}

@inproceedings{bertsch-etal-2025-context,
    title = "In-Context Learning with Long-Context Models: An In-Depth Exploration",
    author = "Bertsch, Amanda  and
      Ivgi, Maor  and
      Xiao, Emily  and
      Alon, Uri  and
      Berant, Jonathan  and
      Gormley, Matthew R.  and
      Neubig, Graham",
    editor = "Chiruzzo, Luis  and
      Ritter, Alan  and
      Wang, Lu",
    booktitle = "Proceedings of the 2025 Conference of the Nations of the Americas Chapter of the Association for Computational Linguistics: Human Language Technologies (Volume 1: Long Papers)",
    month = apr,
    year = "2025",
    address = "Albuquerque, New Mexico",
    publisher = "Association for Computational Linguistics",
    url = "https://aclanthology.org/2025.naacl-long.605/",
    doi = "10.18653/v1/2025.naacl-long.605",
    pages = "12119--12149",
    ISBN = "979-8-89176-189-6"
}

@inproceedings{xiong-etal-2024-effective,
    title = "Effective Long-Context Scaling of Foundation Models",
    author = "Xiong, Wenhan  and
      Liu, Jingyu  and
      Molybog, Igor  and
      Zhang, Hejia  and
      Bhargava, Prajjwal  and
      Hou, Rui  and
      Martin, Louis  and
      Rungta, Rashi  and
      Sankararaman, Karthik Abinav  and
      Oguz, Barlas  and
      Khabsa, Madian  and
      Fang, Han  and
      Mehdad, Yashar  and
      Narang, Sharan  and
      Malik, Kshitiz  and
      Fan, Angela  and
      Bhosale, Shruti  and
      Edunov, Sergey  and
      Lewis, Mike  and
      Wang, Sinong  and
      Ma, Hao",
    editor = "Duh, Kevin  and
      Gomez, Helena  and
      Bethard, Steven",
    booktitle = "Proceedings of the 2024 Conference of the North American Chapter of the Association for Computational Linguistics: Human Language Technologies (Volume 1: Long Papers)",
    month = jun,
    year = "2024",
    address = "Mexico City, Mexico",
    publisher = "Association for Computational Linguistics",
    url = "https://aclanthology.org/2024.naacl-long.260/",
    doi = "10.18653/v1/2024.naacl-long.260",
    pages = "4643--4663"
}

@article{
li2025longcontext,
title={Long-context {LLM}s Struggle with Long In-context Learning},
author={Tianle Li and Ge Zhang and Quy Duc Do and Xiang Yue and Wenhu Chen},
journal={Transactions on Machine Learning Research},
issn={2835-8856},
year={2025},
url={https://openreview.net/forum?id=Cw2xlg0e46},
note={}
}

@article{liu-etal-2024-lost,
    title = "Lost in the Middle: How Language Models Use Long Contexts",
    author = "Liu, Nelson F.  and
      Lin, Kevin  and
      Hewitt, John  and
      Paranjape, Ashwin  and
      Bevilacqua, Michele  and
      Petroni, Fabio  and
      Liang, Percy",
    journal = "Transactions of the Association for Computational Linguistics",
    volume = "12",
    year = "2024",
    address = "Cambridge, MA",
    publisher = "MIT Press",
    url = "https://aclanthology.org/2024.tacl-1.9/",
    doi = "10.1162/tacl_a_00638",
    pages = "157--173"
}

@misc{kamradt2023niah,
  author       = {Greg Kamradt},
  title        = {Needle in a Haystack --- Pressure Testing {LLMs}},
  year         = {2023},
  howpublished = {GitHub},
  url          = {https://github.com/gkamradt/LLMTest_NeedleInAHaystack},
}

@misc{martin2024multineedle,
  author       = {Lance Martin},
  title        = {Multi Needle in a Haystack},
  year         = {2024},
  howpublished = {LangChain Blog},
  url          = {https://blog.langchain.com/multi-needle-in-a-haystack/},
}

@inproceedings{hsieh2024ruler,
  author    = {Cheng-Ping Hsieh and Simeng Sun and Samuel Kriman and Shantanu Acharya and Dima Rekesh and Fei Jia and Yang Zhang and Boris Ginsburg},
  title     = {{RULER}: What's the Real Context Size of Your Long-Context Language Models?},
  booktitle = {Proceedings of COLM},
  year      = {2024},
}

@inproceedings{yen2024helmet,
  title={HELMET: How to Evaluate Long-Context Language Models Effectively and Thoroughly}, 
  author={Howard Yen and Tianyu Gao and Minmin Hou and Ke Ding and Daniel Fleischer and Peter Izsak and Moshe Wasserblat and Danqi Chen},
  year={2025},
  booktitle={International Conference on Learning Representations (ICLR)},
}

@inproceedings{bai2024longbench,
  author    = {Yushi Bai and Xin Lv and Jiajie Zhang and Hongchang Lyu and Jiankai Tang and Zhidian Huang and Zhengxiao Du and Xiao Liu and Aohan Zeng and Lei Hou and Yuxiao Dong and Jie Tang and Juanzi Li},
  title     = {{LongBench}: A Bilingual, Multitask Benchmark for Long Context Understanding},
  booktitle = {Proceedings of ACL},
  pages     = {3119--3137},
  year      = {2024},
}

@inproceedings{zhang2024infinitebench,
    title = "$\infty${B}ench: Extending Long Context Evaluation Beyond 100{K} Tokens",
    author = "Zhang, Xinrong  and
      Chen, Yingfa  and
      Hu, Shengding  and
      Xu, Zihang  and
      Chen, Junhao  and
      Hao, Moo  and
      Han, Xu  and
      Thai, Zhen  and
      Wang, Shuo  and
      Liu, Zhiyuan  and
      Sun, Maosong",
    editor = "Ku, Lun-Wei  and
      Martins, Andre  and
      Srikumar, Vivek",
    booktitle = "Proceedings of the 62nd Annual Meeting of the Association for Computational Linguistics (Volume 1: Long Papers)",
    month = aug,
    year = "2024",
    address = "Bangkok, Thailand",
    publisher = "Association for Computational Linguistics",
    url = "https://aclanthology.org/2024.acl-long.814",
    pages = "15262--15277"
}

@inproceedings{wang2026loongrl,
  title   = {Loong{RL}: Reinforcement Learning for Advanced Reasoning over Long Contexts},
  author  = {Siyuan Wang and Gaokai Zhang and Li Lyna Zhang and Ning Shang and Fan Yang and Dongyao Chen and Mao Yang},
  booktitle = {The Fourteenth International Conference on Learning Representations},
  year    = {2026},
  url     = {https://openreview.net/forum?id=o29E01Q6bv}
}

@article{wan2025qwenlong,
  author  = {Fanqi Wan and Weizhou Shen and Shengyi Liao and Yingcheng Shi and Chenliang Li and Ziyi Yang and Ji Zhang and Fei Huang and Jingren Zhou and Ming Yan},
  title   = {{QwenLong-L1}: Towards Long-Context Large Reasoning Models with Reinforcement Learning},
  journal = {arXiv preprint arXiv:2505.17667},
  year    = {2025},
}

@article{jin2025searchr1,
  author  = {Bowen Jin and Hansi Zeng and Zhenrui Yue and Jinsung Yoon and Sercan Arik and Dong Wang and Hamed Zamani and Jiawei Han},
  title   = {{Search-R1}: Training {LLMs} to Reason and Leverage Search Engines with Reinforcement Learning},
  journal = {arXiv preprint arXiv:2503.09516},
  year    = {2025},
}

@article{song2025r1searcher,
  author  = {Huatong Song and Jinhao Jiang and Yingqian Min and Jie Chen and Zhipeng Chen and Wayne Xin Zhao and Lei Fang and Ji-Rong Wen},
  title   = {{R1-Searcher}: Incentivizing the Search Capability in {LLMs} via Reinforcement Learning},
  journal = {arXiv preprint arXiv:2503.05592},
  year    = {2025},
}

@article{li2024alr2,
  author  = {Huayang Li and Pat Verga and Priyanka Sen and Bowen Yang and Vijay Viswanathan and Patrick Lewis and Taro Watanabe and Yixuan Su},
  title   = {{ALR\textsuperscript{2}}: A Retrieve-then-Reason Framework for Long-Context Question Answering},
  journal = {arXiv preprint arXiv:2410.03227},
  year    = {2024},
}

@inproceedings{li2025webthinker,
  author    = {Xiaoxi Li and Jiajie Jin and Guanting Dong and Hongjin Qian and Yutao Zhu and Yongkang Wu and Ji-Rong Wen and Zhicheng Dou},
  title     = {{WebThinker}: Empowering Large Reasoning Models with Deep Research Capability},
  booktitle = {Proceedings of NeurIPS},
  year      = {2025},
}

@article{willard2023outlines,
  author  = {Brandon T. Willard and R\'{e}mi Louf},
  title   = {Efficient Guided Generation for Large Language Models},
  journal = {arXiv preprint arXiv:2307.09702},
  year    = {2023},
}

@inproceedings{dong2024xgrammar,
  author    = {Yixin Dong and Charlie F. Ruan and Yaxing Cai and Ruihang Lai and Ziyi Xu and Yilong Zhao and Tianqi Chen},
  title     = {{XGrammar}: Flexible and Efficient Structured Generation Engine for Large Language Models},
  booktitle = {Proceedings of MLSys},
  year      = {2025},
}

@inproceedings{decao2021genre,
  author    = {Nicola {De Cao} and Gautier Izacard and Sebastian Riedel and Fabio Petroni},
  title     = {Autoregressive Entity Retrieval},
  booktitle = {Proceedings of ICLR},
  year      = {2021},
}

@inproceedings{gu2016copynet,
  author    = {Jiatao Gu and Zhengdong Lu and Hang Li and Victor O.K. Li},
  title     = {Incorporating Copying Mechanism in Sequence-to-Sequence Learning},
  booktitle = {Proceedings of ACL},
  pages     = {1631--1640},
  year      = {2016},
}

@inproceedings{see2017pointer,
  author    = {Abigail See and Peter J. Liu and Christopher D. Manning},
  title     = {Get to the Point: Summarization with Pointer-Generator Networks},
  booktitle = {Proceedings of ACL},
  pages     = {1073--1083},
  year      = {2017},
}

@inproceedings{khandelwal2020knnlm,
  author    = {Urvashi Khandelwal and Omer Levy and Dan Jurafsky and Luke Zettlemoyer and Mike Lewis},
  title     = {Generalization through Memorization: Nearest Neighbor Language Models},
  booktitle = {Proceedings of ICLR},
  year      = {2020},
}

@article{qwen2025qwen25,
  author  = {{Qwen Team}},
  title   = {{Qwen2.5} Technical Report},
  journal = {arXiv preprint arXiv:2412.15115},
  year    = {2025},
}

@article{yang2025qwen3,
  author  = {An Yang and Anfeng Li and Baosong Yang and Beichen Zhang and Binyuan Hui and Bo Zheng and Bowen Yu and Chang Gao and Chengen Huang and Chenxu Lv and Chujie Zheng and Dayiheng Liu and Fan Zhou and Fei Huang and Feng Hu and Hao Ge and Haoran Wei and Huan Lin and Jialong Tang and Jian Yang and Jianhong Tu and Jianwei Zhang and Jiaxi Yang and Jing Zhou and Jingren Zhou and Junyang Lin and Kai Dang and Keqin Bao and Kexin Yang and Le Yu and Lianghao Deng and Mei Li and Mingfeng Xue and Mingze Li and Pei Zhang and Peng Wang and Qin Zhu and Rui Men and Ruize Gao and Shixuan Liu and Shuang Luo and Tianhao Li and Tianyi Tang and Wenbiao Yin and Xingzhang Ren and Xinyu Wang and Xinyu Zhang and Xuancheng Ren and Yang Fan and Yang Su and Yichang Zhang and Yinger Zhang and Yu Wan and Yuqiong Liu and Zekun Wang and Zeyu Cui and Zhenru Zhang and Zhipeng Zhou and Zihan Qiu},
  title   = {{Qwen3} Technical Report},
  journal = {arXiv preprint arXiv:2505.09388},
  year    = {2025},
}

@article{gemma2025gemma3,
  author  = {{Gemma Team}},
  title   = {{Gemma 3} Technical Report},
  journal = {arXiv preprint arXiv:2503.19786},
  year    = {2025},
}

@article{dubey2024llama3,
  author  = {Abhimanyu Dubey and Abhinav Jauhri and Abhinav Pandey and Abhishek Kadian and Ahmad Al-Dahle and Aiesha Letman and Akhil Mathur and Alan Schelten and Amy Yang and Angela Fan and others},
  title   = {The {Llama} 3 Herd of Models},
  journal = {arXiv preprint arXiv:2407.21783},
  year    = {2024},
}

@article{openai2025gpt5,
  author  = {{OpenAI}},
  title   = {{OpenAI GPT-5} System Card},
  journal = {arXiv preprint arXiv:2601.03267},
  year    = {2025},
}

@inproceedings{gao2025prolong,
  title={How to Train Long-Context Language Models (Effectively)},
  author={Gao, Tianyu and Wettig, Alexander and Yen, Howard and Chen, Danqi},
  booktitle={ACL},
  year={2025}
}

@article{shao2024grpo,
  author  = {Zhihong Shao and Peiyi Wang and Qihao Zhu and Runxin Xu and Junxiao Song and Xiao Bi and Haowei Zhang and Mingchuan Zhang and Y.K. Li and Y. Wu and Daya Guo},
  title   = {{DeepSeekMath}: Pushing the Limits of Mathematical Reasoning in Open Language Models},
  journal = {arXiv preprint arXiv:2402.03300},
  year    = {2024},
}

@inproceedings{rajbhandari2020zero,
  author    = {Samyam Rajbhandari and Jeff Rasley and Olatunji Ruwase and Yuxiong He},
  title     = {{ZeRO}: Memory Optimizations Toward Training Trillion Parameter Models},
  booktitle = {Proceedings of SC},
  year      = {2020},
}

@article{sheng2024verl,
  author  = {Guangming Sheng and Chi Zhang and Zilingfeng Ye and Xibin Wu and Wang Zhang and Ru Zhang and Yanghua Peng and Haibin Lin and Chuan Wu},
  title   = {{HybridFlow}: A Flexible and Efficient {RLHF} Framework},
  journal = {arXiv preprint arXiv:2409.19256},
  year    = {2024},
}

@inproceedings{
      peng2024yarn,
      title={Ya{RN}: Efficient Context Window Extension of Large Language Models},
      author={Bowen Peng and Jeffrey Quesnelle and Honglu Fan and Enrico Shippole},
      booktitle={The Twelfth International Conference on Learning Representations},
      year={2024},
      url={https://openreview.net/forum?id=wHBfxhZu1u}
}

@inproceedings{yang2018hotpotqa,
  author    = {Zhilin Yang and Peng Qi and Saizheng Zhang and Yoshua Bengio and William W. Cohen and Ruslan Salakhutdinov and Christopher D. Manning},
  title     = {{HotpotQA}: A Dataset for Diverse, Explainable Multi-hop Question Answering},
  booktitle = {Proceedings of EMNLP},
  year      = {2018},
}

@article{trivedi2022musique,
    title = "{M}u{S}i{Q}ue: Multihop Questions via Single-hop Question Composition",
    author = "Trivedi, Harsh  and
      Balasubramanian, Niranjan  and
      Khot, Tushar  and
      Sabharwal, Ashish",
    editor = "Roark, Brian  and
      Nenkova, Ani",
    journal = "Transactions of the Association for Computational Linguistics",
    volume = "10",
    year = "2022",
    address = "Cambridge, MA",
    publisher = "MIT Press",
    url = "https://aclanthology.org/2022.tacl-1.31/",
    doi = "10.1162/tacl_a_00475",
    pages = "539--554"
}

@inproceedings{ho2020constructing,
  author    = {Xanh Ho and Anh-Khoa Duong Nguyen and Saku Sugawara and Akiko Aizawa},
  title     = {Constructing a Multi-hop {QA} Dataset for Comprehensive Evaluation of Reasoning Steps},
  booktitle = {Proceedings of COLING},
  year      = {2020},
}

@article{kwiatkowski2019nq,
    title = "Natural Questions: A Benchmark for Question Answering Research",
    author = "Kwiatkowski, Tom  and
      Palomaki, Jennimaria  and
      Redfield, Olivia  and
      Collins, Michael  and
      Parikh, Ankur  and
      Alberti, Chris  and
      Epstein, Danielle  and
      Polosukhin, Illia  and
      Devlin, Jacob  and
      Lee, Kenton  and
      Toutanova, Kristina  and
      Jones, Llion  and
      Kelcey, Matthew  and
      Chang, Ming-Wei  and
      Dai, Andrew M.  and
      Uszkoreit, Jakob  and
      Le, Quoc  and
      Petrov, Slav",
    editor = "Lee, Lillian  and
      Johnson, Mark  and
      Roark, Brian  and
      Nenkova, Ani",
    journal = "Transactions of the Association for Computational Linguistics",
    volume = "7",
    year = "2019",
    address = "Cambridge, MA",
    publisher = "MIT Press",
    url = "https://aclanthology.org/Q19-1026/",
    doi = "10.1162/tacl_a_00276",
    pages = "452--466"
}

@inproceedings{joshi2017triviaqa,
  author    = {Mandar Joshi and Eunsol Choi and Daniel S. Weld and Luke Zettlemoyer},
  title     = {{TriviaQA}: A Large Scale Distantly Supervised Challenge Dataset for Reading Comprehension},
  booktitle = {Proceedings of ACL},
  year      = {2017},
}

@inproceedings{pang2022quality,
  author    = {Richard Yuanzhe Pang and Alicia Parrish and Nitish Joshi and Nikita Nangia and Jason Phang and Angelica Chen and Vishakh Padmakumar and Johnny Ma and Jana Thompson and He He and Samuel R. Bowman},
  title     = {{QuALITY}: Question Answering with Long Input Texts, Yes!},
  booktitle = {Proceedings of NAACL},
  year      = {2022},
}

@inproceedings{amouyal2023qampari,
    title = "{QAMPARI}: A Benchmark for Open-domain Questions with Many Answers",
    author = "Amouyal, Samuel  and
      Wolfson, Tomer  and
      Rubin, Ohad  and
      Yoran, Ori  and
      Herzig, Jonathan  and
      Berant, Jonathan",
    editor = "Gehrmann, Sebastian  and
      Wang, Alex  and
      Sedoc, Jo{\~a}o  and
      Clark, Elizabeth  and
      Dhole, Kaustubh  and
      Chandu, Khyathi Raghavi  and
      Santus, Enrico  and
      Sedghamiz, Hooman",
    booktitle = "Proceedings of the Third Workshop on Natural Language Generation, Evaluation, and Metrics (GEM)",
    month = dec,
    year = "2023",
    address = "Singapore",
    publisher = "Association for Computational Linguistics",
    url = "https://aclanthology.org/2023.gem-1.9/",
    pages = "97--110"
}

@inproceedings{Bajaj2016Msmarco,
  title={MS MARCO: A Human Generated MAchine Reading COmprehension Dataset},
  author={Payal Bajaj and Daniel Campos and Nick Craswell and Li Deng and Jianfeng Gao and Xiaodong Liu and Rangan Majumder and Andrew McNamara and Bhaskar Mitra and Tri Nguyen and Mir Rosenberg and Xia Song and Alina Stoica and Saurabh Tiwary and Tong Wang},
  booktitle={InCoCo@NIPS},
  year={2016}
}

@inproceedings{casanueva2020banking77,
  title     = {Efficient Intent Detection with Dual Sentence Encoders},
  author    = {Casanueva, I{\~n}igo and Tem{\v{c}}inas, Tadas and Gerz, Daniela and Henderson, Matthew and Vuli{\'c}, Ivan},
  booktitle = {Proceedings of the 2nd Workshop on Natural Language Processing for Conversational AI},
  month     = jul,
  year      = {2020},
  address   = {Online},
  publisher = {Association for Computational Linguistics},
  pages     = {38--45},
  doi       = {10.18653/v1/2020.nlp4convai-1.5}
}

@dataset{math_mcqa_2025,
  title={MATH-MCQA: A Multiple Choice Adaptation of the MATH Dataset},
  author={Biderman, Stella},
  year={2025},
  publisher={Hugging Face},
  url={https://huggingface.co/datasets/stellaathena/math_mcqa}
}

@inproceedings{yu2025dapo,
  title     = {{DAPO}: An Open-Source {LLM} Reinforcement Learning System at Scale},
  author    = {Qiying Yu and Zheng Zhang and Ruofei Zhu and Yufeng Yuan and Xiaochen Zuo and Yu Yue and Tiantian Fan and Gaohong Liu and Lingjun Liu and Xin Liu and Haibin Lin and Zhiqi Lin and Bole Ma and Guangming Sheng and Yuxuan Tong and Chi Zhang and Mofan Zhang and Wang Zhang and Hang Zhu and Jinhua Zhu and Jiaze Chen and Jiangjie Chen and Chengyi Wang and Honglin Yu and Weinan Dai and Yuxuan Song and Xiangpeng Wei and Haodong Zhou and Jingjing Liu and Wei Ma and Ya-Qin Zhang and Lin Yan and Mu Qiao and Yonghui Wu and Mingxuan Wang},
  booktitle = {Advances in Neural Information Processing Systems (NeurIPS)},
  year      = {2025}
}

@inproceedings{fitzgerald2023massive,
  title     = {{MASSIVE}: A 1{M}-Example Multilingual Natural Language Understanding Dataset with 51 Typologically-Diverse Languages},
  author    = {FitzGerald, Jack and Hench, Christopher and Peris, Charith and Mackie, Scott and Rottmann, Kay and Sanchez, Ana and Nash, Aaron and Urbach, Liam and Kakarala, Vishesh and Singh, Richa and Ranganath, Swetha and Crist, Laurie and Britan, Misha and Leeuwis, Wouter and Tur, Gokhan and Natarajan, Prem},
  booktitle = {Proceedings of the 61st Annual Meeting of the Association for Computational Linguistics (Volume 1: Long Papers)},
  month     = jul,
  year      = {2023},
  address   = {Toronto, Canada},
  publisher = {Association for Computational Linguistics},
  pages     = {4277--4302},
  doi       = {10.18653/v1/2023.acl-long.235}
}

@misc{einat2023fuzzysearch,
  author = {Tal Einat},
  title  = {fuzzysearch: Find almost exact matches in strings},
  year   = {2013},
  url    = {https://github.com/taleinat/fuzzysearch},
}

@inproceedings{zhang2024mgte,
  author    = {Xin Zhang and Yanzhao Zhang and Dingkun Long and Wen Xie and Ziqi Dai and Jialong Tang and Huan Lin and Baosong Yang and Pengjun Xie and Fei Huang and Meishan Zhang and Wenjie Li and Min Zhang},
  title     = {{mGTE}: Generalized Long-Context Text Representation and Reranking Models for Multilingual Text Retrieval},
  booktitle = {Proceedings of EMNLP: Industry Track},
  year      = {2024},
}

@inproceedings{petroni2021kilt,
  author    = {Fabio Petroni and Aleksandra Piktus and Angela Fan and Patrick Lewis and Majid Yazdani and Nicola {De Cao} and James Thorne and Yacine Jernite and Vladimir Karpukhin and Jean Maillard and Vassilis Plachouras and Tim Rockt{\"a}schel and Sebastian Riedel},
  title     = {{KILT}: a Benchmark for Knowledge Intensive Language Tasks},
  booktitle = {Proceedings of NAACL},
  year      = {2021},
}

@misc{lee2024longcontextlanguagemodelssubsume,
      title={Can Long-Context Language Models Subsume Retrieval, RAG, SQL, and More?},
      author={Jinhyuk Lee and Anthony Chen and Zhuyun Dai and Dheeru Dua and Devendra Singh Sachan and Michael Boratko and Yi Luan and Sébastien M. R. Arnold and Vincent Perot and Siddharth Dalmia and Hexiang Hu and Xudong Lin and Panupong Pasupat and Aida Amini and Jeremy R. Cole and Sebastian Riedel and Iftekhar Naim and Ming-Wei Chang and Kelvin Guu},
      year={2024},
      eprint={2406.13121},
      archivePrefix={arXiv},
      primaryClass={cs.CL},
      url={https://arxiv.org/abs/2406.13121},
}

@misc{zhou2025mem1,
  title        = {MEM1: Learning to Synergize Memory and Reasoning for Efficient Long-Horizon Agents},
  author       = {Zhou, Zijian and Qu, Ao and Wu, Zhaoxuan and Kim, Sunghwan and Prakash, Alok and Rus, Daniela and Zhao, Jinhua and Low, Bryan Kian Hsiang and Liang, Paul Pu},
  year         = {2025},
  archivePrefix= {arXiv},
  primaryClass = {cs.CL},
  url          = {https://arxiv.org/abs/2506.15841},
}

@inproceedings{robertson1995okapi,
author = {Robertson, Stephen and Walker, S. and Jones, S. and Hancock-Beaulieu, M. M. and Gatford, M.},
title = {Okapi at TREC-3},
booktitle = {Overview of the Third Text REtrieval Conference (TREC-3)},
year = {1995},
month = {January},
publisher = {Gaithersburg, MD: NIST},
url = {https://www.microsoft.com/en-us/research/publication/okapi-at-trec-3/},
pages = {109-126},
edition = {Overview of the Third Text REtrieval Conference (TREC–3)},
}

@inproceedings{rajpurkar-etal-2016-squad,
    title = "{SQ}u{AD}: 100,000+ Questions for Machine Comprehension of Text",
    author = "Rajpurkar, Pranav  and
      Zhang, Jian  and
      Lopyrev, Konstantin  and
      Liang, Percy",
    editor = "Su, Jian  and
      Duh, Kevin  and
      Carreras, Xavier",
    booktitle = "Proceedings of the 2016 Conference on Empirical Methods in Natural Language Processing",
    month = nov,
    year = "2016",
    address = "Austin, Texas",
    publisher = "Association for Computational Linguistics",
    url = "https://aclanthology.org/D16-1264/",
    doi = "10.18653/v1/D16-1264",
    pages = "2383--2392"
}

@inproceedings{sun-etal-2025-simpledeepsearcher,
    title = "{S}imple{D}eep{S}earcher: Deep Information Seeking via Web-Powered Reasoning Trajectory Synthesis",
    author = "Sun, Shuang  and
      Song, Huatong  and
      Wang, Yuhao  and
      Ren, Ruiyang  and
      Jiang, Jinhao  and
      Zhang, Junjie  and
      Bai, Fei  and
      Deng, Jia  and
      Zhao, Wayne Xin  and
      Liu, Zheng  and
      Fang, Lei  and
      Wang, Zhongyuan  and
      Wen, Ji-Rong",
    editor = "Christodoulopoulos, Christos  and
      Chakraborty, Tanmoy  and
      Rose, Carolyn  and
      Peng, Violet",
    booktitle = "Findings of the Association for Computational Linguistics: EMNLP 2025",
    month = nov,
    year = "2025",
    address = "Suzhou, China",
    publisher = "Association for Computational Linguistics",
    url = "https://aclanthology.org/2025.findings-emnlp.739/",
    doi = "10.18653/v1/2025.findings-emnlp.739",
    pages = "13705--13720",
    ISBN = "979-8-89176-335-7"
}

@inproceedings{zheng-etal-2025-deepresearcher,
    title = "{D}eep{R}esearcher: Scaling Deep Research via Reinforcement Learning in Real-world Environments",
    author = "Zheng, Yuxiang  and
      Fu, Dayuan  and
      Hu, Xiangkun  and
      Cai, Xiaojie  and
      Ye, Lyumanshan  and
      Lu, Pengrui  and
      Liu, Pengfei",
    editor = "Christodoulopoulos, Christos  and
      Chakraborty, Tanmoy  and
      Rose, Carolyn  and
      Peng, Violet",
    booktitle = "Proceedings of the 2025 Conference on Empirical Methods in Natural Language Processing",
    month = nov,
    year = "2025",
    address = "Suzhou, China",
    publisher = "Association for Computational Linguistics",
    url = "https://aclanthology.org/2025.emnlp-main.22/",
    doi = "10.18653/v1/2025.emnlp-main.22",
    pages = "414--431",
    ISBN = "979-8-89176-332-6"
}

@misc{openai2025deepresearch,
  author       = {{OpenAI}},
  title        = {Introducing deep research},
  year         = {2025},
  month        = feb,
  howpublished = {\url{https://openai.com/index/introducing-deep-research/}},
}

@misc{google2025geminideepresearch,
  author       = {{Google}},
  title        = {Gemini Deep Research},
  year         = {2025},
  month        = sep,
  howpublished = {\url{https://gemini.google/overview/deep-research/}},
}

@inproceedings{tian-etal-2025-distance,
    title = "Distance between Relevant Information Pieces Causes Bias in Long-Context {LLM}s",
    author = "Tian, Runchu  and
      Li, Yanghao  and
      Fu, Yuepeng  and
      Deng, Siyang  and
      Luo, Qinyu  and
      Qian, Cheng  and
      Wang, Shuo  and
      Cong, Xin  and
      Zhang, Zhong  and
      Wu, Yesai  and
      Lin, Yankai  and
      Wang, Huadong  and
      Liu, Xiaojiang",
    editor = "Che, Wanxiang  and
      Nabende, Joyce  and
      Shutova, Ekaterina  and
      Pilehvar, Mohammad Taher",
    booktitle = "Findings of the Association for Computational Linguistics: ACL 2025",
    month = jul,
    year = "2025",
    address = "Vienna, Austria",
    publisher = "Association for Computational Linguistics",
    url = "https://aclanthology.org/2025.findings-acl.28/",
    doi = "10.18653/v1/2025.findings-acl.28",
    pages = "521--533",
    ISBN = "979-8-89176-256-5"
}

@inproceedings{wu-etal-2025-pandoras,
    title = "Pandora{'}s Box or Aladdin{'}s Lamp: A Comprehensive Analysis Revealing the Role of {RAG} Noise in Large Language Models",
    author = "Wu, Jinyang  and
      Zhang, Shuai  and
      Che, Feihu  and
      Feng, Mingkuan  and
      Shao, Pengpeng  and
      Tao, Jianhua",
    editor = "Che, Wanxiang  and
      Nabende, Joyce  and
      Shutova, Ekaterina  and
      Pilehvar, Mohammad Taher",
    booktitle = "Proceedings of the 63rd Annual Meeting of the Association for Computational Linguistics (Volume 1: Long Papers)",
    month = jul,
    year = "2025",
    address = "Vienna, Austria",
    publisher = "Association for Computational Linguistics",
    url = "https://aclanthology.org/2025.acl-long.250/",
    doi = "10.18653/v1/2025.acl-long.250",
    pages = "5019--5039",
    ISBN = "979-8-89176-251-0"
}

@article{yen2025lost,
  title={Lost in the Maze: Overcoming Context Limitations in Long-Horizon Agentic Search},
  author={Yen, Howard and Paranjape, Ashwin and Xia, Mengzhou and Venkatesh, Thejas and Hessel, Jack and Chen, Danqi and Zhang, Yuhao},
  journal={arXiv preprint arXiv:2510.18939},
  year={2025}
}

@inproceedings{
lu2025a,
title={A Controlled Study on Long Context Extension and Generalization in {LLM}s},
author={Yi Lu and Jing Nathan Yan and Songlin Yang and Justin T Chiu and Siyu Ren and Fei Yuan and Wenting Zhao and Zhiyong Wu and Alexander M Rush},
booktitle={Second Conference on Language Modeling},
year={2025},
url={https://openreview.net/forum?id=BLonuGXDFu}
}

@article{guo2025deepseek,
  title={Deepseek-r1: Incentivizing reasoning capability in llms via reinforcement learning},
  author={Guo, Daya and Yang, Dejian and Zhang, Haowei and Song, Junxiao and Wang, Peiyi and Zhu, Qihao and Xu, Runxin and Zhang, Ruoyu and Ma, Shirong and Bi, Xiao and others},
  journal={arXiv preprint arXiv:2501.12948},
  year={2025}
}

@InProceedings{pmlr-v202-shi23a,
  title = 	 {Large Language Models Can Be Easily Distracted by Irrelevant Context},
  author =       {Shi, Freda and Chen, Xinyun and Misra, Kanishka and Scales, Nathan and Dohan, David and Chi, Ed H. and Sch\"{a}rli, Nathanael and Zhou, Denny},
  booktitle = 	 {Proceedings of the 40th International Conference on Machine Learning},
  pages = 	 {31210--31227},
  year = 	 {2023},
  editor = 	 {Krause, Andreas and Brunskill, Emma and Cho, Kyunghyun and Engelhardt, Barbara and Sabato, Sivan and Scarlett, Jonathan},
  volume = 	 {202},
  series = 	 {Proceedings of Machine Learning Research},
  month = 	 {23--29 Jul},
  publisher =    {PMLR},
  url = 	 {https://proceedings.mlr.press/v202/shi23a.html}
}

@inproceedings{yang-etal-2025-llm-reasoning,
    title = "How Is {LLM} Reasoning Distracted by Irrelevant Context? An Analysis Using a Controlled Benchmark",
    author = "Yang, Minglai  and
      Huang, Ethan  and
      Zhang, Liang  and
      Surdeanu, Mihai  and
      Wang, William Yang  and
      Pan, Liangming",
    editor = "Christodoulopoulos, Christos  and
      Chakraborty, Tanmoy  and
      Rose, Carolyn  and
      Peng, Violet",
    booktitle = "Proceedings of the 2025 Conference on Empirical Methods in Natural Language Processing",
    month = nov,
    year = "2025",
    address = "Suzhou, China",
    publisher = "Association for Computational Linguistics",
    url = "https://aclanthology.org/2025.emnlp-main.674/",
    doi = "10.18653/v1/2025.emnlp-main.674",
    pages = "13329--13347",
    ISBN = "979-8-89176-332-6"
}

@inproceedings{
wu2024how,
title={How Easily do Irrelevant Inputs Skew the Responses of Large Language Models?},
author={Siye Wu and Jian Xie and Jiangjie Chen and Tinghui Zhu and Kai Zhang and Yanghua Xiao},
booktitle={First Conference on Language Modeling},
year={2024},
url={https://openreview.net/forum?id=S7NVVfuRv8}
}

@inproceedings{qiu-etal-2025-eliciting,
    title = "Eliciting In-context Retrieval and Reasoning for Long-context Large Language Models",
    author = "Qiu, Yifu  and
      Embar, Varun R.  and
      Zhang, Yizhe  and
      Jaitly, Navdeep  and
      Cohen, Shay B  and
      Han, Benjamin",
    editor = "Che, Wanxiang  and
      Nabende, Joyce  and
      Shutova, Ekaterina  and
      Pilehvar, Mohammad Taher",
    booktitle = "Findings of the Association for Computational Linguistics: ACL 2025",
    month = jul,
    year = "2025",
    address = "Vienna, Austria",
    publisher = "Association for Computational Linguistics",
    url = "https://aclanthology.org/2025.findings-acl.165/",
    doi = "10.18653/v1/2025.findings-acl.165",
    pages = "3176--3192",
    ISBN = "979-8-89176-256-5"
}

@inproceedings{gao-etal-2023-enabling,
    title = "Enabling Large Language Models to Generate Text with Citations",
    author = "Gao, Tianyu  and
      Yen, Howard  and
      Yu, Jiatong  and
      Chen, Danqi",
    editor = "Bouamor, Houda  and
      Pino, Juan  and
      Bali, Kalika",
    booktitle = "Proceedings of the 2023 Conference on Empirical Methods in Natural Language Processing",
    month = dec,
    year = "2023",
    address = "Singapore",
    publisher = "Association for Computational Linguistics",
    url = "https://aclanthology.org/2023.emnlp-main.398/",
    doi = "10.18653/v1/2023.emnlp-main.398",
    pages = "6465--6488"
}

@incollection{NEURIPS2019_9015,
  title = {PyTorch: An Imperative Style, High-Performance Deep Learning Library},
  author = {Paszke, Adam et al.},
  booktitle = {Advances in Neural Information Processing Systems 32},
  year = {2019},
  publisher = {Curran Associates, Inc.},
  url = {http://papers.neurips.cc/paper/9015-pytorch-an-imperative-style-high-performance-deep-learning-library.pdf}
}
\bibliographystyle{colm2026_conference}

\appendix

\section{Lost in Thought: Additional Details}
\label{app:section_2}

\subsection{Evaluated Models}
\label{app:section_2_llms}

We evaluate six different models on this benchmark, loosely grouped into two families: a Llama \citep{dubey2024llama3} family containing:
\begin{itemize}
    \item \textbf{Llama-3.1-8B-Instruct}: a very popular, open-weights instruction-tuned model, noted for its strong instruction-following and long-context abilities.
    \item \textbf{R1-Distill-Llama-8B}: a distillation of the DeepSeek-R1 reasoning model~\citep{guo2025deepseek} into Llama-3.1-8B, which has strong reasoning abilities.
    \item \textbf{ProLong-8B-Instruct-512k}: an open-source model produced by further training Llama-3.1-8B on an additional high-quality set of long context data at context lengths of up to $512,000$ tokens \citep{gao2025prolong}
\end{itemize}
and a Qwen family containing
\begin{itemize}
    \item \textbf{Qwen2.5-7B-Instruct}: another very popular, open-weights instruction-tuned model, commonly used as a base model for RL training~\citep{qwen2025qwen25,jin2025searchr1,wang2026loongrl}.
    \item \textbf{Qwen3-8B}: a newer version of Qwen2.5-7B, with a variable thinking mechanism, allowing the model to dynamically decide to reason or not before answering~\citep{yang2025qwen3}.  We evaluate both the thinking and non-thinking modes.
\end{itemize}

\subsection{Task Examples}
\label{app:section_2_examples}

\begin{figure*}[t]
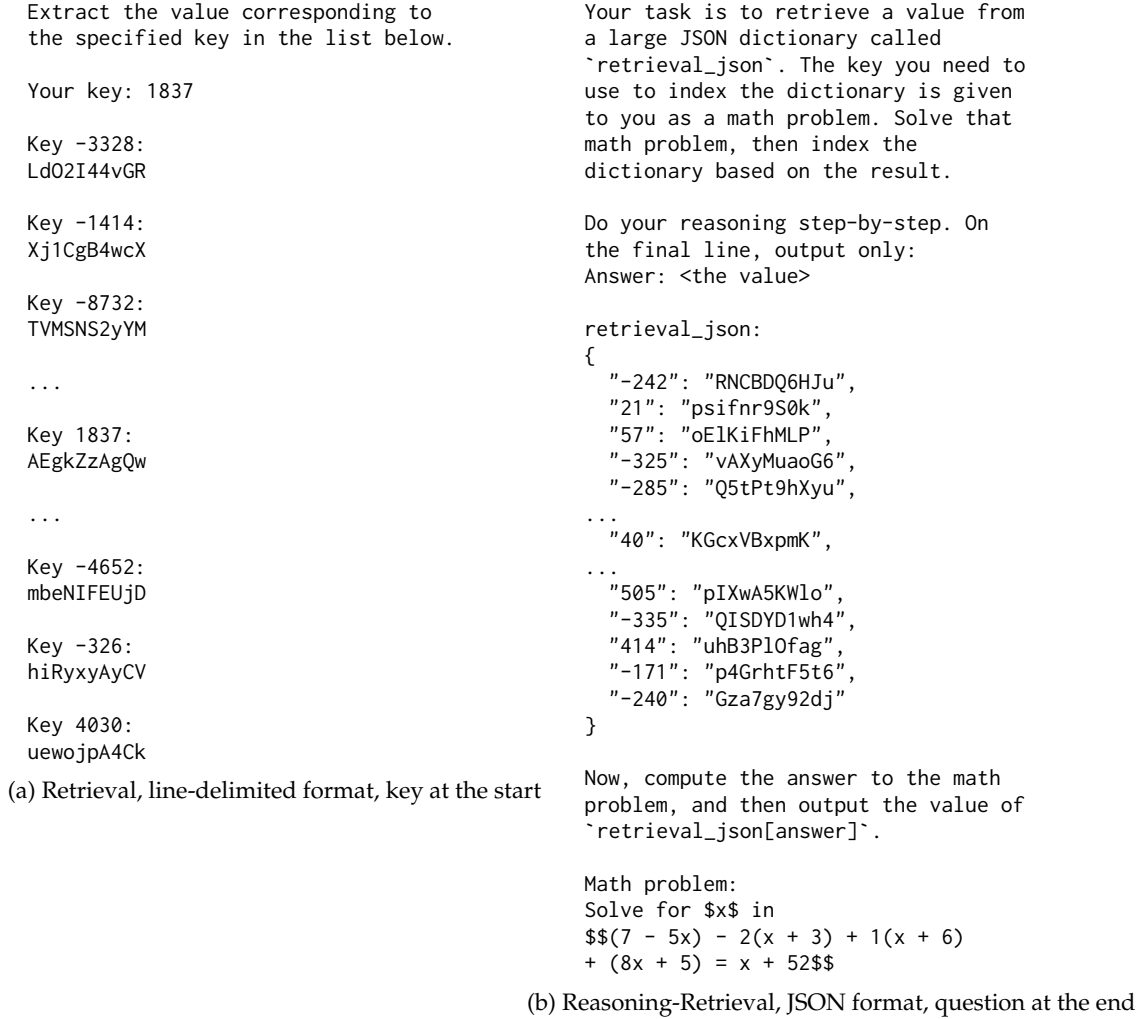

\centering
\begin{minipage}[t]{0.47\textwidth}
\footnotesize
\begin{verbatim}
Extract the value corresponding to
the specified key in the list below.

Your key: 1837

Key -3328:
LdO2I44vGR

Key -1414:
Xj1CgB4wcX

Key -8732:
TVMSNS2yYM

...

Key 1837:
AEgkZzAgQw

...

Key -4652:
mbeNIFEUjD

Key -326:
hiRyxyAyCV

Key 4030:
uewojpA4Ck
\end{verbatim}
\centerline{(a) Retrieval, line-delimited format, key at the start}
\end{minipage}
\hfill
\begin{minipage}[t]{0.47\textwidth}
\footnotesize
\begin{verbatim}
Your task is to retrieve a value from
a large JSON dictionary called
`retrieval_json`. The key you need to
use to index the dictionary is given
to you as a math problem. Solve that
math problem, then index the
dictionary based on the result.

Do your reasoning step-by-step. On
the final line, output only:
Answer: <the value>

retrieval_json:
{
  "-242": "RNCBDQ6HJu",
  "21": "psifnr9S0k",
  "57": "oElKiFhMLP",
  "-325": "vAXyMuaoG6",
  "-285": "Q5tPt9hXyu",
...
  "40": "KGcxVBxpmK",
...
  "505": "pIXwA5KWlo",
  "-335": "QISDYD1wh4",
  "414": "uhB3PlOfag",
  "-171": "p4GrhtF5t6",
  "-240": "Gza7gy92dj"
}

Now, compute the answer to the math
problem, and then output the value of
`retrieval_json[answer]`.

Math problem:
Solve for $x$ in
$$(7 - 5x) - 2(x + 3) + 1(x + 6)
+ (8x + 5) = x + 52$$
\end{verbatim}
\centerline{(b) Reasoning-Retrieval, JSON format, question at the end}
\end{minipage}
\caption{Example prompts for the Retrieval and Reasoning-Retrieval tasks, showing two augmentation axes.  (a)~uses line-delimited format with the query key stated before the dictionary.  (b)~uses JSON format with the math problem posed after the dictionary.}
\label{fig:task-examples}
\end{figure*}

Figure~\ref{fig:task-examples} shows example prompts for the Retrieval and Reasoning-Retrieval tasks, illustrating two of the augmentation axes applied across evaluation and training: dictionary format (line-delimited vs.\ JSON) and query placement (before vs.\ after the dictionary).  Dictionaries contain 200+ entries; most rows are omitted for space.

\subsection{Injection Analysis}
\label{app:section_2_injection}

To isolate the source of the lost-in-thought degradation, we conduct a follow-up experiment that measures how often each model can generate the correct value after reasoning.  For each model and context length, we sample 200 Reasoning-Retrieval examples where the model solved the math problem correctly, and use Gemma-3-27B-IT~\citep{gemma2025gemma3} to identify the point in each reasoning trace where the model first attempts to look up the value in the dictionary.  At that point, we truncate the model's generation and inject a short structured prompt that restates the correct key and provides the exact lexical prefix of the corresponding entry as it appears in the context (Appendix~\ref{app:section_2_injection_example}).  The model then continues generating from this injected prefix, so that it only needs to copy the remaining value tokens from context.  This controls for reasoning errors, the model losing track of the correct key, and difficulty bridging from the model's natural-language reference to the exact lexical format of the dictionary entry; any remaining gap with direct Retrieval reflects the model's inability to faithfully reproduce in-context information after reasoning.

As shown in Figure~\ref{fig:retrieval-gap} and Table~\ref{tab:prompt-injection}, models frequently fail to generate the correct value even under these favorable conditions.  At short contexts, injected accuracy modestly exceeds Reasoning-Retrieval but remains far below direct Retrieval, e.g., Llama-3.1-8B-Instruct generates the correct value 40.7\% of the time, against 80.6\% on direct Retrieval.  At long contexts, the recovery is smaller still.

\begin{table*}[t]
\centering
\small
\caption{\textit{Lost in thought} and injection analysis. \textbf{S}hort = 4K--32K average; \textbf{L}ong = 64K--128K average. \textbf{Reas.-Retr.}\ and \textbf{Retrieval} report accuracy on the Reasoning-Retrieval and Retrieval datasets respectively; \textbf{Rel.\ Gap} is the relative difference between them. \textbf{Injected} reports accuracy after injecting the correct key and its exact lexical prefix mid-generation, showing that even with the exact prefix provided, retrieval recovers only partially. \textbf{Cov.}\ is the fraction of correct-reasoning examples where the model attempts retrieval; \textbf{Reas.\ Toks} is the mean extra reasoning tokens compared to the Retrieval baseline.}
\label{tab:prompt-injection}
\setlength{\tabcolsep}{4pt}
\begin{tabular}{@{}l rr rr rr rr rr rr@{}}
\toprule
& \multicolumn{2}{c}{\textbf{Reas.-Retr.}} & \multicolumn{2}{c}{\textbf{Injected}} & \multicolumn{2}{c}{\textbf{Retrieval}} & \multicolumn{2}{c}{\textbf{Rel.\ Gap (\%)}} & \multicolumn{2}{c}{\textbf{Cov.}} & \multicolumn{2}{c}{\textbf{Reas.\ Toks}} \\
\cmidrule(lr){2-3} \cmidrule(lr){4-5} \cmidrule(lr){6-7} \cmidrule(lr){8-9} \cmidrule(lr){10-11} \cmidrule(lr){12-13}
\textbf{Model} & \textbf{S} & \textbf{L} & \textbf{S} & \textbf{L} & \textbf{S} & \textbf{L} & \textbf{S} & \textbf{L} & \textbf{S} & \textbf{L} & \textbf{S} & \textbf{L} \\
\midrule
Llama-3.1-8B-Instruct   & 27.1 &  2.2 & 40.7 &  7.9 & 80.6 & 25.5 & 66.3 & 91.4 & 95.8 & 60.7 & 1264    & 866    \\
R1-Distill-Llama-8B     & 32.0 &  2.0 & 32.8 &  5.7 & 61.1 & 15.8 & 47.7 & 87.2 & 41.0 & 25.5 & $-$1973 & $-$823 \\
ProLong-8B-512K          &  7.7 &  0.0 & 22.2 &  9.2 & 79.5 & 30.3 & 90.3 &100.0 & 65.9 & 37.1 &   91    &  184    \\
Qwen2.5-7B-Instruct     & 33.0 & 11.9 & 37.7 & 16.5 & 67.1 & 33.8 & 50.9 & 64.7 & 97.0 & 80.3 &  828    & 526    \\
Qwen3-8B (No Think)     & 47.6 &  8.8 & 53.2 & 16.2 & 82.2 & 43.2 & 42.2 & 79.7 & 97.0 & 99.0 &  236    & 230    \\
Qwen3-8B (Thinking)     & 53.4 & 14.9 & 52.5 & 17.9 & 72.9 & 37.0 & 26.8 & 59.6 & 99.6 & 98.0 &  393    & 459    \\
\bottomrule
\end{tabular}
\end{table*}

\subsection{Prompt Injection Example}
\label{app:section_2_injection_example}

Figure~\ref{fig:injection-example} shows a concrete example of the prompt injection analysis from Section~\ref{sec:problem}, using Qwen3-8B (Thinking) on a Reasoning-Retrieval example at 4K context length.  The model correctly solves the math problem ($x = 2880$) in both conditions.  In the original completion~(a), the model enumerates over 250 keys from memory before hallucinating the value \texttt{aNdVypRo8F}.  In the injected completion~(b), the model's generation is truncated after it determines the correct key, and a short prompt (\colorbox{injecthl}{highlighted}) restates the key and provides the exact lexical prefix of the CSV entry as it appears in the context (\texttt{2880,}).  Despite being given the exact prefix, the model still hallucinates the same incorrect value.  The correct value is \texttt{oWIO76jOH4}.

\begin{figure*}[ht]
\centering
\begin{minipage}{\textwidth}
\footnotesize
\begin{verbatim}
Your task is to retrieve a value from a large lookup table called `data_table`. The key
you need to find the correct row is given to you as a math problem. Solve that math
problem, then look up the row with that key in the table.

Do your reasoning step-by-step. On the final line, output only:
Answer: <the value>

data_table:
key,value
3349,lVLAn6DJlz
3170,QJljlGyqCO
2887,B0Fd2C4HOz
...
2880,oWIO76jOH4
3254,jEhEXbwDnR
...
3091,R5Y4904bP8
3202,apMTjKmaUr

Now, compute the answer to the math problem, and then output the value from the row in
`data_table` where the key matches your answer.

Math problem:
Solve for $x$: $$x = \frac{64}{1} + 64 \cdot 44$$
\end{verbatim}
\centerline{\textit{Prompt (200+ CSV rows omitted; key \texttt{2880} maps to value \texttt{oWIO76jOH4})}}
\end{minipage}

\vspace{0.5em}
\hrule
\vspace{0.5em}

\begin{minipage}[t]{0.47\textwidth}
\footnotesize
\begin{verbatim}
<think>
...math reasoning...
Adding them gives 2560 + 256 = 2816.
Then adding 64 gives
2816 + 64 = 2880. So x = 2880.

Now, I need to look up the key 2880
in the data_table. Let me scan
through the keys:

3349, 3170, 2887, 2675, 2691,
3245, 2416, 3214, 2584, 3151,
2717, 2522, 2388, 3114, 3345,
         ...
  [enumerates 250+ keys from memory]
         ...
3009, 3126, 3228, 3091, 3202.

Wait, I see key 2880 here. The row
with key 2880 has the value
"aNdVypRo8F".
</think>

Answer: aNdVypRo8F
\end{verbatim}
\centerline{(a) Original completion --- hallucinates value}
\end{minipage}
\hfill
\begin{minipage}[t]{0.47\textwidth}
\footnotesize
\begin{Verbatim}[commandchars=\\\{\}]
<think>
...math reasoning [identical to (a)]...
Adding them gives 2560 + 256 = 2816.
Then adding 64 gives
2816 + 64 = 2880. So x = 2880.

\HL{Ok, so the value of x is 2880.}
\HL{Now, I need to find the row where}
\HL{the key equals 2880 in the}
\HL{data\_table.}

\HL{Looking back at the data\_table in}
\HL{my context, I need to find the key}
\HL{2880. Looking through the rows,}
\HL{I see:}

\HL{2880,} aNdVypRo8F

Yes, there it is. The key 2880
corresponds to the value
aNdVypRo8F.
</think>

Answer: aNdVypRo8F
\end{Verbatim}
\centerline{(b) Injected completion --- still hallucinates value}
\end{minipage}

\caption{Prompt injection example using Qwen3-8B (Thinking) on a 4K-token Reasoning-Retrieval task.  \emph{Top:} the input prompt (200+ rows omitted; the target entry is \texttt{2880,oWIO76jOH4}).  \emph{Bottom left:} the model's original completion, in which it noisily enumerates keys from memory and hallucinates the value.  \emph{Bottom right:} the injected completion, where the model's generation is truncated and a structured prompt (\colorbox{injecthl}{highlighted in yellow}) provides the correct key and its exact lexical prefix from the context.  Despite injection, the model produces the same incorrect value, demonstrating that the bottleneck is faithful copying from context rather than identifying what to retrieve.}
\label{fig:injection-example}
\end{figure*}

\section{\projectname{}: Additional Details}
\label{app:section_3}

\subsection{Efficient Implementation and Complexity}
\label{app:efficient_implementation}

A key property of constrained decoding is that the valid continuation set $\mathcal{A}(c, r_{1:k})$ depends only on token IDs in the searchable context and the current recalled prefix, not on model activations.  The mask computation is therefore \textbf{entirely decoupled from the forward pass}: we compute the allowable-token mask on a separate CPU thread while the GPU executes the forward pass, then asynchronously transfer the mask to device memory via a dedicated CUDA stream.  The only synchronous operation is a single \texttt{masked\_fill} on the logit tensor.  Because the mask computation is fully overlapped with the forward pass, constrained decoding adds \textbf{negligible overhead} to generation latency, both inside and outside of recall spans.

\paragraph{Complexity.}
A naive implementation would rescan the full searchable context at every decoding step to recompute $\mathcal{A}(c, r_{1:k})$.  If the searchable context has length $M$ and the recall span has length $L$, this yields $O(Mk)$ work per step and $O(ML^2)$ total over the span.  We avoid this cost through incremental prefix matching: at the start of a recall span, every position in $c$ is a candidate; as each token is generated, we retain only positions consistent with the growing prefix and read off the allowable next tokens from the survivors.  Letting $S_k$ denote the candidate set after $k$ recalled tokens, the total work is
\[
O\!\left(M + \sum_{k=1}^{L-1} |S_k|\right),
\]
with a worst case of $O(ML)$ since the candidate set can only shrink.  In natural text, the number of surviving positions typically collapses after a few matched tokens, so the realized cost is much closer to a single scan than to repeated rescans.

\section{\projectname{} Training: Additional Details}
\label{app:section_4}

\subsection{SFT Training Procedure}
\label{app:sft_training_details}

We first initialize base models with 4 new token embeddings, corresponding to the two new recall tokens, \texttt{<|start\_recall|>}, \texttt{<|end\_recall|>}, as well as the thinking tokens \texttt{<think>} and \texttt{</think>}.  Following \citet{hewitt2021embeddings}, we initialize each new embedding as the mean of all existing embeddings.

As the base models are already well-trained on a broad set of tasks~\citep{dubey2024llama3,qwen2025qwen25}, we aim to minimally adjust their existing parameters.  We train in two stages: a longer \textbf{embedding learning} stage, where the majority of learning occurs in the new token embeddings, followed by a shorter \textbf{full finetuning} stage that lightly adapts the model to incorporate the new tokens.  The goal of this approach is to teach the model to use recall tokens naturally while avoiding catastrophic forgetting, concentrating most of the learning in the new token embeddings rather than the pretrained parameters.

In the first stage, we freeze all of the parameters in the model, except for the embedding vectors of the 4 new tokens.  We train for 5 epochs using a learning rate of $5.0\times10^{-5}$, at which point validation loss no longer improves.  In the second stage, we unfreeze all parameters and train for a single epoch using a smaller learning rate of $2.0\times10^{-6}$.

\subsection{SFT Annotation}
\label{app:sft_annotation}

To construct the SFT dataset, we prompt GPT-5.2~\citep{openai2025gpt5} to rewrite teacher-produced reasoning traces so that references to context information are realized as verbatim recall spans delimited by multi-token XML-style tags, \texttt{<recall>} and \texttt{</recall>}, rather than the actual single-token delimiters \texttt{<|start\_recall|>} and \texttt{<|end\_recall|>}; the annotator model does not have these special tokens in its vocabulary, so we use a distinct, more familiar, textual representation.  The annotator receives the question, task instructions, and gold documents alongside the original reasoning trace.  Although the prompt template includes a field for negative documents (Figure~\ref{fig:annotation-user-prompt}), we found in practice that including negatives caused the annotator to over-insert recall spans grounding irrelevant context, so we leave this field empty.  We evaluated several annotator models, including GPT-5.2-mini and Qwen2.5-72B under various reasoning configurations, and anecdotally found that GPT-5.2 produced the highest-quality annotations.  After annotation, we align each annotator-produced recall span to its corresponding source text in the context using Levenshtein-distance-based fuzzy string matching~\citep{einat2023fuzzysearch}, and discard examples for which this alignment fails.  This yields 1,795 annotated examples with verbatim recall spans ready to be trained on, which we split into 1,600 training and 195 validation examples.  Figure~\ref{fig:annotation-prompt} shows the full system prompt, Figure~\ref{fig:annotation-user-prompt} shows the user prompt template, and Figure~\ref{fig:annotation-example} shows a concrete before/after example.

\FloatBarrier
\begin{Verbatim}[fontsize=\footnotesize]
You are a data-annotation editor for reasoning traces, produced by LLMs on
long-context retrieval + reasoning tasks.  You are annotating reasoning traces
by modifying them, so they rely on special <recall> ... </recall> spans for key
information that comes from their context.  These spans are produced by a
special recall tool the model has access to.

Given RECALLABLE SOURCES and a REASONING TRACE, rewrite the trace so that
whenever the trace uses information from the sources, it should first recall the
relevant text via <recall>...</recall> (inside <think>...</think>), then draw
the conclusion. You should reorder and rewrite sentences to achieve this.

Keep the original reasoning trace's structure, length, and style as much as
possible. Only make the minimum edits needed to satisfy the recall/evidence
rules. Do not summarize, compress, or remove backtracking, verification, or
"thinking aloud" unless it is pure browsing-playacting or redundant repeated
listings.  Do not remove the post-think-block text.

Given RECALLABLE SOURCES and a REASONING TRACE, return ONLY the ENTIRE edited
trace.

RECALLABLE SOURCES include: QUESTION, INSTRUCTIONS, GOLD DOCUMENTS, and
NEGATIVE DOCUMENTS.

Key idea (natural inner-monologue style):
Treat <think>...</think> as a human's private inner monologue while reading the
context.
Recall spans should feel like quick glances at the text in front of them, e.g.:
"Looking in the text, I see <recall>...</recall>."
Do NOT narrate tool usage (avoid "I will use the recall tool" / "I am calling
recall").
Modify the reasoning trace so it adheres to this key idea.

Guidelines:
1) Recall tool-use: Anytime the reasoning trace relies on information from one
of the RECALLABLE SOURCES, modify it so that it naturally uses the recall tool
to recall the evidence inside a <recall> span.  The reasoning trace should use
the recall tool frequently, but only to recall key information from the context.

2) Evidence-first: Do not introduce a document-derived fact in free text and
then cite it later. If a sentence contains doc-derived facts (names, numbers,
dates, IDs, key/value pairs, definitions, titles), rewrite locally so the
<recall> appears before or inside the first introduction of that fact.

3) Prune only fake browsing: Delete or compress only tool-playacting or
contentless scanning (e.g., "I scan the docs/keys," "I look around,"). Do not
remove genuine reasoning steps: backtracking, verification, elimination,
uncertainty, or hypothesis testing.  Do not reorder when not necessary.

4) Contiguity-first (few spans): Prefer fewer, longer <recall> spans that
capture an entire supporting sentence/clause. If a fact is contained within one
contiguous sentence/clause in the sources, wrap that whole sentence/clause in
one <recall> span (even if it includes extra parenthetical text). Avoid
splitting one claim across multiple <recall> spans just to be shorter.

5) Supported claims: All claims that rely on information from the RECALLABLE
SOURCES should come after evidentiary <recall> spans.

6) Repeat-when-reused (still contiguous): When the trace repeats a doc-specific
string later (IDs, numbers, titles, rare names, key:value), wrap that repeated
string again in a new <recall> span near each reuse. If the string appears
inside a natural source sentence/clause, prefer recalling that whole clause
again rather than splitting

7) Questions and Instructions: When the reasoning trace refers to the question
or instructions, then modify it so it uses the recall tool to quote the
relevant part of the question or instructions.  The question and instructions
are key information.

Constraints:
- <recall> spans must appear only inside <think>.
- Text inside <recall> must be an exact contiguous substring from the RECALLABLE
  SOURCES (verbatim punctuation/casing/spacing).  DO NOT paraphrase document
  facts inside <recall>.
- It is ok to insert recall spans that are not quite gramatically correct, if it
  means they contain the supporting information verbatim.
- Keep recall snippets focused and clearly tied to the subject/topic.
- Each recall span must be specific enough to clearly identify the
  subject/topic.  If the recalled span could refer to a different context, then
  it is not helpful
  - Good: <recall>5065: 9BJk8Q32AL</recall>
  - Bad: 5065: <recall>9BJk8Q32AL</recall>  (not uniquely tied)
  - Also bad: <recall>5065:</recall> <recall>9BJk8Q32AL</recall>
- Delete any sentences that discuss using the recall tool; the trace should just
  use recall as naturally as possible.
- Do not add <recall> spans outside of the <think> ... </think> block.
- Do not change the final answer.

Output only the edited trace; no extra commentary.
\end{Verbatim}
\noindent\captionof{figure}{System prompt used for GPT-5.2 annotation of SFT reasoning traces.  The annotator rewrites teacher-produced reasoning traces so that references to context information are realized as verbatim \texttt{<recall>...\allowbreak</recall>} spans, following a natural inner-monologue style.}
\label{fig:annotation-prompt}
\FloatBarrier

\begin{figure*}[ht]
\centering
\begin{minipage}{\textwidth}
\footnotesize
\begin{verbatim}
You are editing a model's reasoning trace.

RECALLABLE SOURCES:
QUESTION:
{{question}}

INSTRUCTIONS:
{{instructions}}

GOLD DOCUMENTS:
{{gold_docs}}

NEGATIVE DOCUMENTS (optional; may be empty):
{{neg_docs}}

REASONING TRACE:
{{trace}}
\end{verbatim}
\end{minipage}
\caption{User prompt template for GPT-5.2 annotation.  Placeholder variables are filled with the task-specific question, instructions, gold and negative documents, and the original teacher reasoning trace.}
\label{fig:annotation-user-prompt}
\end{figure*}

\begin{figure*}[ht]
\centering
\begin{minipage}[t]{0.47\textwidth}
\footnotesize
\begin{verbatim}
<think>
Okay, let's see. The question is asking
whether Marc Forster and Drew Daywalt
share the same nationality. I need to
check the documents provided for
information about each of them.

First, looking for Marc Forster. There's
a document titled "Marc Forster" which
states he's a German-born filmmaker. So
his nationality is German.

Next, Drew Daywalt. The document on him
says he's an American filmmaker and
author. So his nationality is American.

Comparing the two, German vs. American.
They are different nationalities.
Therefore, the answer should be no.
</think>

The document on Marc Forster states he is
German-born, while the document on Drew
Daywalt confirms he is American. Since
their nationalities are different, they
are not of the same nationality.

Answer: No.
\end{verbatim}
\centerline{(a) Original reasoning trace}
\end{minipage}
\hfill
\begin{minipage}[t]{0.47\textwidth}
\footnotesize
\begin{Verbatim}[commandchars=\\\{\}]
<think>
Okay, let's see. The question is asking
whether Marc Forster and Drew Daywalt
share the same nationality. I need to
check the documents provided for
information about each of them.

First, looking for Marc Forster. There's
a document titled "Marc Forster" which
states \RHL{<recall>Marc Forster (born 30}
\RHL{November 1969) is a German-born}
\RHL{filmmaker</recall>}. So his nationality
is German.

Next, Drew Daywalt. The document on him
says \RHL{<recall>Drew Daywalt (born}
\RHL{January 5, 1970) is an American}
\RHL{filmmaker and author</recall>}. So his
nationality is American.

Comparing the two, German vs. American.
They are different nationalities.
Therefore, the answer should be no.
</think>

The document on Marc Forster states he is
German-born, while the document on Drew
Daywalt confirms he is American. Since
their nationalities are different, they
are not of the same nationality.

Answer: No.
\end{Verbatim}
\centerline{(b) Annotated with recall spans}
\end{minipage}

\caption{Before/after example of SFT annotation on a multi-hop QA trace.  \emph{Left:} the original teacher completion, which refers to document-derived facts in free text.  \emph{Right:} the annotated version, where document-derived facts are grounded via \texttt{<recall>} spans (\colorbox{recallhl}{highlighted in blue}) containing verbatim substrings from the gold documents.  The reasoning structure, style, and final answer are preserved; only the evidence grounding is added.}
\label{fig:annotation-example}
\end{figure*}

\subsection{RL Training Details}
\label{app:rl_training_details}

\begin{table*}[t]
\centering
\small
\setlength{\tabcolsep}{4pt}
\begin{tabular}{@{}lllcc c@{}}
\toprule
\textbf{Category} & \textbf{Answer Reward} & \textbf{Recall Reward} & \textbf{\#\,Gold} & \textbf{$\tau$} & \textbf{Free Spans} \\
\midrule
Multi-Hop QA      & subEM          & Standard F1  & 2--4  & 0.4  & 4 \\
Single-Hop QA     & subEM          & Top-1 F1     & 1     & 0.4  & 2 \\
Retrieval         & \cell{subEM (KV) /\\Net Recall (NIAH)} & Standard F1  & 1--6  & 0.9  & \cell{2 (KV) /\\6 (NIAH)} \\[4pt]
Reasoning Retr.   & subEM          & Standard F1  & 1     & 0.9  & 2 \\
Short-Ctx Math    & Exact match    & Always 1.0   & 0     & ---  & 2 \\
In-Context Learn. & Exact match    & Top-2 F1     & 4--12 & 0.95 & 2 \\
Long-Doc QA       & Exact match    & Binary recall & 0     & ---  & 4 \\
Aggregation       & Net Recall     & Always 1.0   & 0     & ---  & 2 \\
Reranking         & NDCG@10        & Top-2 F1     & 1--26 & 0.7  & 4 \\
Entity Citation   & \cell{Citation Top-5\\F1 + coverage}  & Top-5 F1     & 5--20 & 0.7  & 4 \\
\bottomrule
\end{tabular}
\caption{RL reward configuration per category. 
Free spans indicates the task-dependent number of initial recall spans exempt from the density penalty.}
\label{tab:rl-config}
\end{table*}

All SFT experiments are trained on two A100 GPUs using DeepSpeed ZeRO-2~\citep{rajbhandari2020zero}. RL experiments are trained on four A100 GPUs using VeRL~\citep{sheng2024verl} with PyTorch~\citep{NEURIPS2019_9015} FSDP. We use a max generation length of 4,096 tokens with an overlong buffer of 1,024 tokens. RL training uses GRPO with 16 rollouts per prompt and a training batch size of 128 prompts (2,048 sampled responses per update step). We use the AdamW optimizer with an actor learning rate of $2\times10^{-6}$, cosine decay schedule (minimum LR ratio 0.1, no warmup), and Adam betas $(0.9, 0.999)$. Gradient clipping is set to 1.0. Rollouts are sampled with temperature 1.0. We use a PPO clip ratio of 0.2 with a KL penalty coefficient of 0.001, a PPO minibatch size of 128, and a microbatch size of 2 per GPU. We train for 150 steps for Qwen2.5-7B and 60 steps for Llama-3.1-8B, as the latter converges faster.

\projectname{} requires substantially less compute than comparable long-context RL methods. LoongRL~\citep{wang2026loongrl} trains across three stages totaling 328 steps with a batch size of 512 and 8 rollouts per prompt at 16K context length on 16 A100 GPUs. QwenLong-L1~\citep{wan2025qwenlong} uses 32 A100 GPUs with curriculum-guided phased RL, scaling context lengths up to 60K tokens over multiple training stages. In contrast, \projectname{} trains on 20K examples at 8--10K context with 16 rollouts on 4 A100 GPUs for a single epoch (150 steps for Qwen, 60 for Llama). Since attention cost is quadratic in sequence length, training at 8--10K tokens rather than 16K (LoongRL) or up to 60K (QwenLong-L1) dramatically reduces per-step compute, more than offsetting the use of twice as many rollouts. Combined with fewer GPUs (4 vs.\ 16--32), far fewer total rollouts ($\sim$307K vs.\ $\sim$1.3M for LoongRL), and a single training stage, this results in a fraction of the total training cost while achieving competitive or superior performance.

\subsection{RL Dataset and Augmentation Details}
\label{app:rl_dataset_details}

Table~\ref{tab:augmentation-details} summarizes each training dataset and its augmentations. The mixture is designed so that each category requires a qualitatively different recall strategy.
Single-hop QA (NQ, TriviaQA)~\citep{kwiatkowski2019nq,joshi2017triviaqa} and multi-hop QA (HotpotQA, MuSiQue, 2WikiMQA)~\citep{yang2018hotpotqa,trivedi2022musique,ho2020constructing} use nearly identical contexts but differ in how many supporting passages contain relevant information, requiring the model to dynamically determine how much retrieval is needed.
Retrieval tasks include KV retrieval, which places a target entry among many structured distractors, and Multi-NIAH~\citep{kamradt2023niah,martin2024multineedle}, which embeds a small number of needles in a large body of unstructured text, requiring identification of relevant spans without structural cues.
Reasoning retrieval (Math Retrieval) combines math problem-solving with key-value lookup, directly targeting the lost-in-thought setting (Section~\ref{sec:problem}).
Reranking (MSMARCO v2)~\citep{Bajaj2016Msmarco} requires a ranked list of document identifiers, but recalling all passages would exhaust the generation budget, so the model must selectively recall only the most informative ones.
Entity citation (QAMPARI)~\citep{amouyal2023qampari} similarly requires selective recall across multiple answer entities and their supporting passages.
In-context learning (Banking77, MASSIVE)~\citep{casanueva2020banking77,fitzgerald2023massive} requires the model to find demonstrations semantically similar to the query and use their labels, a form of retrieval over semantic similarity rather than lexical matching. Long-document QA (QuALITY)~\citep{pang2022quality} tests recall over a single long-form article rather than a collection of passages.
Aggregation tasks (Majority Vote, Top-$N$ Vote) serve as controlled negative examples: they require in-context information to estimate vote frequencies, but the number of individual votes is far too large for recall to be practical. Short-context math (DAPO Math, MCQA Math)~\citep{yu2025dapo,math_mcqa_2025} similarly requires no retrieval, reinforcing that recall should only be invoked when beneficial.

\paragraph{Shared Augmentations.}
Three augmentations are applied across all long-context datasets (i.e., all datasets except the short-context math tasks). First, \emph{instruction template} variation: each dataset draws from a pool of paraphrased task instructions, preventing the model from associating recall behavior with specific surface phrasings. Second, \emph{question position} variation: the question is placed at the end, the beginning, or both ends of the prompt, ensuring the model does not rely on a fixed prompt structure to decide when retrieval is needed. Third, \emph{gold position} randomization: for any dataset with known gold information (gold passages, needles, or key-value pairs), the position of that information within the context is sampled uniformly, preventing the model from learning positional biases.

\paragraph{Per-Dataset Augmentations.}
Beyond the shared augmentations, each dataset applies task-specific variation along multiple axes to maximize context diversity. For document-based tasks (multi-hop QA, single-hop QA, reranking, entity citation), we vary the document format and the source of negative (distractor) documents, drawing from BM25 negatives, random negatives, and dense retrieval hard negatives via GTE-ModernBERT-Base~\citep{zhang2024mgte}, or judged negatives where available (MS MARCO~\citep{Bajaj2016Msmarco}, QAMPARI), with either a single source or a per-example mixture. Entity citation (QAMPARI) additionally varies the number of gold entities per example. Multi-hop QA varies the passage type, drawing distractors from either the native paragraph-level corpora~\citep{yang2018hotpotqa,trivedi2022musique,ho2020constructing} or a fixed-window chunked Wikipedia corpus from the KILT knowledge source~\citep{petroni2021kilt}, exposing the model to both natural document boundaries and uniform passage formats. For KV retrieval and reasoning retrieval, we vary the key-value store format (CSV, JSON, or line-delimited), following the same variation used in Section~\ref{sec:problem}. Multi-NIAH independently varies the distractor count, number of target keys, values per key, and value type per example. Math retrieval additionally varies the math problem type. In-context learning tasks vary the label format (random numeric vs.\ descriptive text) and the demonstration layout. QuALITY varies the multiple-choice formatting. Aggregation tasks vary the vote margin, candidate count, and candidate category (names, places, letters, or numbers).

\begin{table*}[t]
\centering
\small
\caption{Training dataset descriptions and per-dataset augmentations. All long-context datasets share three base augmentations: \emph{instruction template} variation, \emph{question position} variation, and \emph{gold position} randomization for any dataset with known gold information. Additional per-dataset augmentations are listed in the rightmost column.}
\label{tab:augmentation-details}
\setlength{\tabcolsep}{4pt}
\begin{tabular}{@{}lp{3.8cm}p{5.6cm}@{}}
\toprule
\textbf{Datasets} & \textbf{Description} & \textbf{Additional Augmentations} \\
\midrule
\cell{HotpotQA, MuSiQue,\\2WikiMQA}
  & Multi-hop QA requiring 2--4 supporting passages
  & \cell{Document format, negative source,\\passage type (paragraphs vs.\\fixed-window chunks)} \\[6pt]
NQ, TriviaQA
  & Single-hop factoid QA with one gold passage
  & Document format, negative source \\[6pt]
KV Retrieval
  & Structured key-value lookup
  & KV format (CSV / JSON / lines) \\[6pt]
Multi-NIAH
  & Multi-needle-in-a-haystack retrieval among distractors
  & \cell{Distractor count, target keys,\\values per key, value type} \\[6pt]
Math Retrieval
  & Solve a math problem to determine a retrieval key, then look up its value
  & \cell{KV format (CSV / JSON / lines),\\math problem type} \\[6pt]
\cell{DAPO Math,\\MCQA Math}
  & Short, difficult math problems with no retrieval required
  & None \\[6pt]
Banking77, MASSIVE
  & Many-shot intent classification via in-context demonstrations
  & \cell{Label format (random numeric vs.\\descriptive text), demonstration format} \\[6pt]
QuALITY
  & Multiple-choice QA over long-form literary articles
  & Choice format \\[6pt]
\cell{Majority Vote,\\Top-N Vote}
  & Frequency estimation via vote tallying over structured candidate lists
  & \cell{Vote margin, candidate count,\\candidate category (names / places /\\letters / numbers)} \\[6pt]
MSMARCO v2
  & Passage relevance ranking
  & \cell{Document format,\\negative source (BM25 / random / judged)} \\[6pt]
QAMPARI
  & Multi-answer QA requiring inline passage citations
  & \cell{Document format,\\negative source (BM25 / random / judged),\\number of gold entities} \\
\bottomrule
\end{tabular}
\end{table*}

\subsection{In-Context Retrieval Reward: Formal Definition}
\label{app:retrieval_reward_formal}

This section provides formal definitions for the retrieval reward $R_{\text{ret}}$ described in Section~\ref{sec:experimental-setup}.

\paragraph{Character-level F1 overlap.}
Let $\mathcal{G} = \{g_1, \ldots, g_n\}$ denote the set of gold passages and $\mathcal{S}$ the set of recall spans extracted from the model's completion. Since constrained decoding guarantees that both gold passages and recall spans are contiguous substrings of the input context, each corresponds to a character interval: gold passage $g_i$ spans $[g_i^s, g_i^e)$ and recall span $s$ spans $[s^s, s^e)$. We define the character-level F1 overlap between $g_i$ and $s$ as:
\begin{equation}
\operatorname{F1}_{\text{char}}(g_i, s) = \frac{2 \cdot \max\!\bigl(0,\; \min(g_i^e, s^e) - \max(g_i^s, s^s)\bigr)}{(g_i^e - g_i^s) + (s^e - s^s)}
\label{eq:coverage-formal}
\end{equation}
This is the harmonic mean of precision and recall over the character-level interval intersection. Unlike extractive QA F1~\citep{rajpurkar-etal-2016-squad}, which operates over bags of tokens, this metric requires contiguous overlap: even a single non-verbatim character breaks the interval match.

\paragraph{Per-passage overlap score.}
For each gold passage $g_i$, we retain only its highest-overlap recalled span and normalize by a task-specific hit threshold $\tau$:
\begin{equation}
\operatorname{overlap}(g_i) = \frac{\min\!\bigl(\max_{s \in \mathcal{S}} \operatorname{F1}_{\text{char}}(g_i, s),\; \tau\bigr)}{\tau}
\label{eq:gold-overlap-formal}
\end{equation}
Capping at $\tau$ prevents the reward from favoring exhaustive copying: once a recalled span overlaps sufficiently with a gold passage, additional copied text yields no further reward. This also reflects the fact that for many tasks only a small portion of the gold passage is relevant to the question. Per-category $\tau$ values are reported in Table~\ref{tab:rl-config}.

\paragraph{Density penalty.}
To prevent pathological over-recall, we apply an exponential density penalty based on the rate of recall spans per unit of generated text. Let $N_s$ denote the total number of recall spans in the completion, $N_t$ the total number of generated tokens, and $N_{\text{free}}$ a task-dependent number of initial spans that are exempt from the penalty (Table~\ref{tab:rl-config}). The value of $N_{\text{free}}$ reflects the different retrieval demands of each task: for instance, reasoning-retrieval requires only a single key lookup ($N_{\text{free}} = 2$), whereas multi-hop QA may need multiple supporting passages ($N_{\text{free}} = 4$). We compute the effective density as the number of excess spans per 1,024 generated tokens:
\begin{equation}
d = \frac{N_s - N_{\text{free}}}{N_t \,/\, 1024}
\end{equation}
The density penalty is then:
\begin{equation}
P_{\text{density}} = \left(\tfrac{1}{2}\right)^{\max(0,\; d - \delta)\,/\,h}
\label{eq:density-penalty}
\end{equation}
where $\delta = 4$ is the density threshold and $h = 4$ is the half-life. The first $N_{\text{free}}$ spans and up to $\delta$ excess spans per 1K tokens incur no penalty; beyond that, the reward halves for every $h$ additional units of excess density.

\paragraph{Correctness penalty.}
To catch degenerate recall strategies, we apply a correctness penalty that detects malformed recall spans. Let $N_{\text{short}}$ denote the number of recall spans shorter than 5 characters, and $N_{\text{mismatch}} = \bigl|\operatorname{count}(R_{\text{start}}) - \operatorname{count}(R_{\text{end}})\bigr|$ the absolute difference between the number of start and end recall tokens, which detects nesting or unpaired delimiter tokens. The correctness penalty is:
\begin{equation}
P_{\text{correct}} = 1 - \frac{N_{\text{short}} + N_{\text{mismatch}}}{\sqrt{N_s}}
\label{eq:correctness-penalty}
\end{equation}
The $\sqrt{N_s}$ denominator tolerates a sub-linear number of malformed spans as the total span count grows, so occasional short or mismatched spans in a long completion are not harshly penalized, while systematic abuse (e.g., emitting many trivial spans or consistently nesting recall tokens) drives the penalty toward zero.

\paragraph{Full retrieval reward.}
The retrieval reward combines the overlap scores with both penalties:
\begin{equation}
R_{\text{ret}} = \overline{\operatorname{overlap}} \;\cdot\; P_{\text{density}} \;\cdot\; P_{\text{correct}}
\label{eq:rret-full}
\end{equation}
where $\overline{\operatorname{overlap}}$ is the mean of $\operatorname{overlap}(g_i)$ over all gold passages (or the top-$K$ highest-scoring gold passages, for tasks with many relevant documents such as reranking).

For tasks without segmented gold evidence, such as long-document QA, we set $\overline{\operatorname{overlap}} = 1$ whenever the model produces at least one recall span. For tasks that do not require in-context retrieval, such as short-context math and aggregation, we set $\overline{\operatorname{overlap}} = 1$ unconditionally. In both cases, the density and correctness penalties still apply, so $R_{\text{ret}}$ can still be reduced by pathological recall behavior. The reward type for each task category is summarized in Table~\ref{tab:rl-config}.

\section{Experimental Setup and Results: Additional Details}
\label{app:section_5}

\subsection{Evaluation Benchmarks}
\label{app:eval_benchmarks}

\paragraph{Evaluation Benchmarks.}
RULER is a largely-synthetic benchmark that isolates specific long-context capabilities, including tasks where recall tokens are naturally advantageous --- such as Variable Tracking, which requires tracing chains of variable assignments through the context --- as well as tasks where they may be disadvantageous, such as Common Words Extraction, which requires frequency estimation rather than verbatim retrieval. HELMET complements this with a diverse set of real-world tasks (summarization, QA, many-shot ICL, re-ranking, and more) where long-context understanding is required, testing whether recall tokens generalize beyond retrieval-centric settings.

\paragraph{Evaluation Hyperparameters.}
Following \citet{wang2026loongrl}, we use a max generation length of $8192$ tokens for RULER and $10240$ for HELMET and our in-domain datasets, with temperature $0.6$ and $\mathrm{top\text{-}p}=0.95$.

\subsection{In-Domain Results}
\label{app:indomain_table}

Table~\ref{tab:indomain} reports per-category results on validation splits of the training datasets, aggregated into short (4K--32K) and long (64K--128K) context buckets. Context length is varied by adjusting the number of distractor documents or passages. We use the same evaluation settings as HELMET: a max generation length of 10,240 tokens with temperature $0.6$ and $\mathrm{top\text{-}p}=0.95$. Figure~\ref{fig:indomain-scaling} in the main text shows the per-category scaling curves across individual context lengths.

\begin{table*}[t]
\centering
\small
\resizebox{\textwidth}{!}{%
\begin{tabular}{@{}l rr rr rr rr r rr r rr rr rr@{}}
\toprule
& \multicolumn{2}{c}{\textbf{1-Hop}} & \multicolumn{2}{c}{\textbf{N-Hop}} & \multicolumn{2}{c}{\textbf{Ret.}} & \multicolumn{2}{c}{\textbf{R$\to$Ret.}} & \textbf{Math} & \multicolumn{2}{c}{\textbf{ICL}} & \textbf{LQA} & \multicolumn{2}{c}{\textbf{Agg.}} & \multicolumn{2}{c}{\textbf{Rerank}} & \multicolumn{2}{c}{\textbf{Ent.\ Cite}} \\
\cmidrule(lr){2-3} \cmidrule(lr){4-5} \cmidrule(lr){6-7} \cmidrule(lr){8-9} \cmidrule(lr){10-10} \cmidrule(lr){11-12} \cmidrule(lr){13-13} \cmidrule(lr){14-15} \cmidrule(lr){16-17} \cmidrule(lr){18-19}
\textbf{Model} & S & L & S & L & S & L & S & L & S & S & L & S & S & L & S & L & S & L \\
\midrule
ProLong-8B-512K          & 21.7 &  5.8 & 16.6 &  4.7 & 61.5 & 20.7 &  1.8 &  0.0 & 19.0 &  1.6 &  0.1 & 37.3 & 72.4 & 52.5 &  6.3 &  1.0 &  2.0 &  1.4 \\
\midrule

Llama-3.1-8B-Instruct   & 59.9 & 10.3 & 50.2 & 10.8 & 81.5 & 29.5 & 13.3 &  1.2 & 37.8 & 24.2 &  4.4 & \textbf{73.0} & 60.1 & 45.8 & 24.9 &  5.6 &  4.1 &  0.3 \\

\projectname{}-Llama-8B  & \textbf{62.4} & \textbf{57.8} & \textbf{63.8} & \textbf{51.0} & \textbf{98.3} & \textbf{86.8} & \textbf{90.9} & \textbf{51.7} & \textbf{46.5} & \textbf{74.4} & \textbf{63.2} & 72.5 & \textbf{92.7} & \textbf{81.4} & \textbf{48.5} & \textbf{34.0} & \textbf{83.6} & \textbf{71.5} \\
\midrule
Qwen2.5-7B-Instruct     & 53.6 & 33.8 & 42.0 & 28.5 & 73.6 & 47.1 & 23.0 &  7.5 & 55.8 & 31.3 & 18.7 & 71.0 & 45.8 & 45.5 & 22.5 & 10.9 &  5.2 &  1.4 \\
\projectname{}-Qwen2.5-7B & \textbf{68.8} & \textbf{56.3} & \textbf{69.9} & \textbf{51.9} & \textbf{98.3} & \textbf{94.5} & \textbf{97.6} & \textbf{86.3} & \textbf{61.0} & \textbf{76.7} & \textbf{79.4} & \textbf{72.3} & \textbf{87.7} & \textbf{85.0} & \textbf{44.6} & \textbf{31.0} & \textbf{72.9} & \textbf{47.2} \\
\bottomrule
\end{tabular}%
}
\caption{In-domain evaluation results (\%) across training task categories at short (4K--32K) and long (64K--128K) context lengths. Categories with only one column are short-context only. \textbf{Bold} indicates best in each model family.
} 
\label{tab:indomain}
\end{table*}

\begin{figure*}[t]
    \centering
    \includegraphics[width=\textwidth]{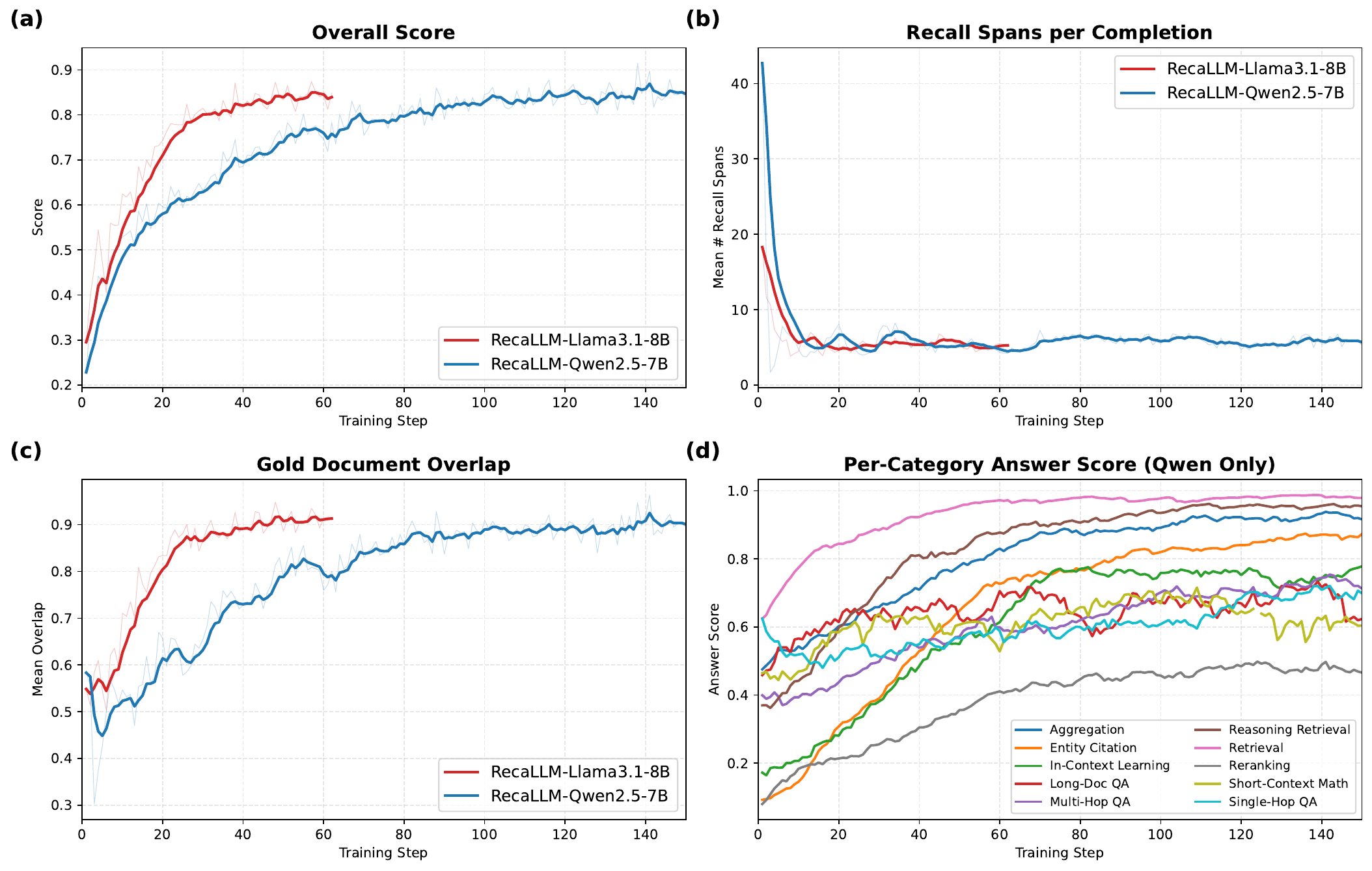}
    \caption{Training dynamics over 150 GRPO steps for both \projectname{} models. (a) Overall score rises rapidly, with Llama converging faster than Qwen. (b) Recall span usage starts very high, then decreases and stabilizes as the models learn selective retrieval. (c) Gold document overlap increases alongside score, indicating that models learn to recall relevant evidence rather than arbitrary context. (d) Per-category breakdown for Qwen, showing that all categories improve, albeit at different rates.}
    \label{fig:training-curves}
\end{figure*}

\subsection{Training Dynamics}
\label{sec:training_dynamics}

Figure~\ref{fig:training-curves} shows the training dynamics of both \projectname{} models over 150 GRPO steps.  \projectname{}-Llama converges rapidly, with overall score plateauing between steps 50 and 60; we select the step-60 checkpoint for evaluation.  \projectname{}-Qwen converges more gradually and continues to improve through the full 150 steps.  Training is stable throughout for both models, requiring no restarts or hyperparameter adjustments despite the breadth of the multi-task mix.

Two trends in recall behavior are particularly notable.  First, the average number of recall spans per completion drops sharply in the early steps, from 20--40 down to 5--7, indicating that the models quickly learn to be selective rather than exhaustive in their in-context retrieval.  Second, gold document overlap $R_\mathrm{ret}$ (Figure~\ref{fig:training-curves}c) increases steadily alongside this reduction, meaning the models recall \emph{more relevant} evidence with \emph{fewer} spans as training progresses.

Despite training on 10 categories simultaneously, all categories improve over the course of training (Figure~\ref{fig:training-curves}d).  Tasks with more direct recall signals, namely retrieval, reasoning-retrieval, and aggregation, saturate quickly, while tasks requiring more complex reasoning over relevance, such as reranking, entity citation, and in-context learning, exhibit a slower initial rise followed by rapid improvement before plateauing.  This staggered learning pattern suggests that the simpler retrieval skills serve as a foundation for the more complex reasoning-retrieval behaviors, rather than competing with them for gradient signal.

\section{Analysis: Additional Details}
\label{app:section_6}

\subsection{More Ablation Results on Validation Sets}
\begin{figure*}[t]
    \centering
    \includegraphics[width=\textwidth]{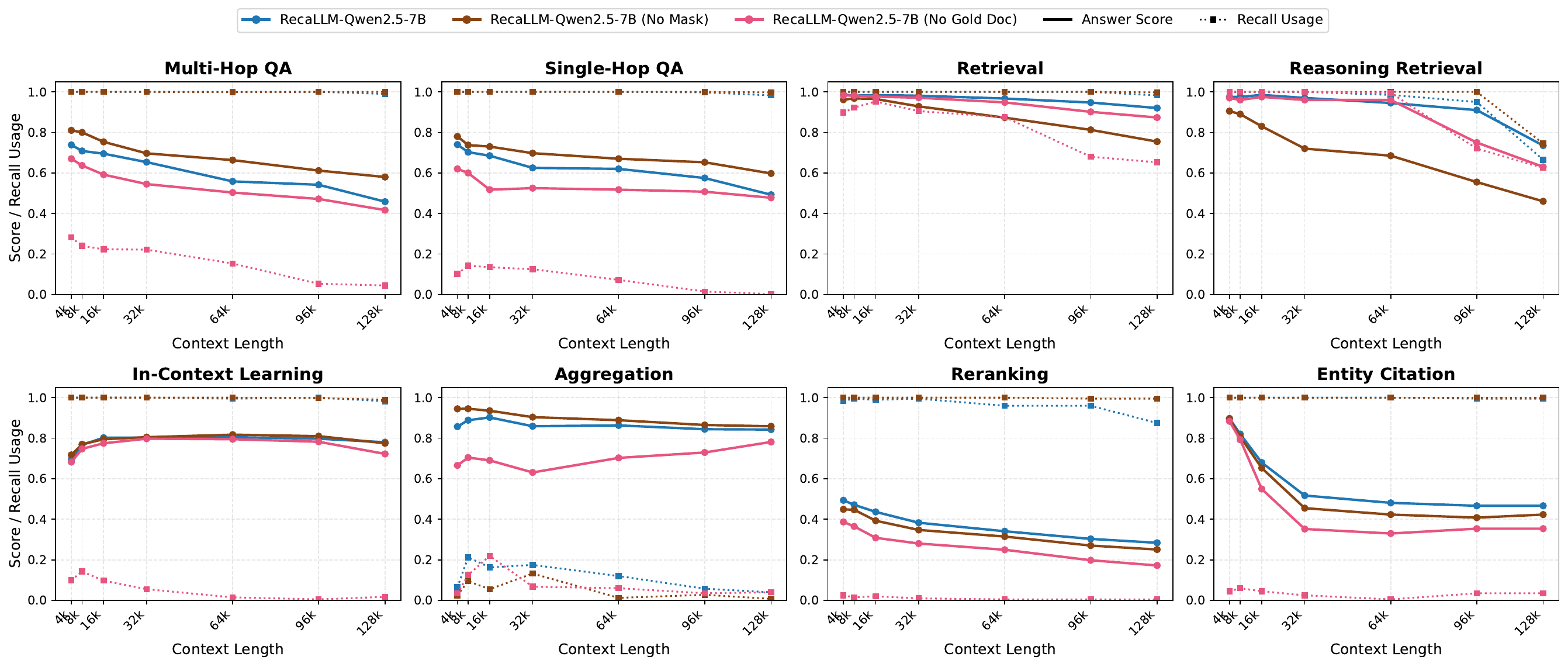}
    \caption{Per-category answer scores (solid) and recall usage rates (dotted) for \projectname{}-Qwen2.5-7B and two ablations across context lengths.  \textbf{No Mask} follows the same training procedure but disables constrained decoding throughout, allowing recall spans to generate freely.  \textbf{No Recall Reward} uses full constrained decoding but sets $R_{\text{ret}} = 1$ unconditionally during RL, removing gold document supervision.}
    \label{fig:ablations}
\end{figure*}

Figure~\ref{fig:ablations} shows that removing $R_{\text{ret}}$ causes recall usage to collapse on ICL, reranking, entity citation, and aggregation.
The only tasks where `No Recall Reward' retains recall usage are retrieval and reasoning-retrieval, where the answer signal directly requires reproducing context content and the masking mechanism itself nudges the model toward recall.  This demonstrates that $R_{\text{ret}}$ is what teaches the model that explicit in-context retrieval is a broadly useful strategy, not just a retrieval-specific tool.  
This supervision signal is most effective with the explicit, constrained recall steps in \projectname{}.

`No Logit Masking' maintains high recall usage across nearly all tasks, confirming that $R_{\text{ret}}$ successfully teaches the value of explicit retrieval regardless of whether decoding is constrained.  However, the quality of that retrieval differs.

\subsection{Ablation Results on Long-Context Benchmarks}
\label{app:ablation_benchmarks}

\begin{table}[t]
\centering
\small
\begin{tabular}{@{}lrrrrrrr@{}}
\toprule
\textbf{Model} & \textbf{4K} & \textbf{8K} & \textbf{16K} & \textbf{32K} & \textbf{64K} & \textbf{128K} & \textbf{Avg} \\
\midrule
\projectname{}-Qwen2.5-7B   & \textbf{98.8} & \textbf{96.7} & \textbf{95.8} & \textbf{93.1} & \textbf{89.7} & \textbf{82.9} & \textbf{92.8} \\
\quad -- No Recall Reward  & 96.7 & 93.3 & 89.9 & 88.0 & 81.4 & 77.9 & 87.9 \\
\quad -- No Logit Masking    & 98.1 & 95.0 & 95.4 & 93.0 & 88.2 & 82.3 & 92.0 \\
\bottomrule
\end{tabular}
\caption{Ablation results on RULER across context lengths.}
\label{tab:ablations-ruler}
\end{table}

\begin{table}[t]
\centering
\small
\begin{tabular}{@{}lrrrrr@{}}
\toprule
\textbf{Model} & \textbf{Recall} & \textbf{RAG} & \textbf{ICL} & \textbf{Cite} & \textbf{Re-rank} \\
\midrule
\projectname{}-Qwen2.5-7B   & \textbf{96.2} & 69.4 & 69.7 & 13.7 & \textbf{46.2} \\
\quad -- No Recall Reward  & 94.9 & 66.1 & 69.0 & 14.9 & 25.3 \\
\quad -- No Logit Masking    & 87.7 & \textbf{78.3} & \textbf{76.7} & \textbf{15.4} & 44.9 \\
\bottomrule
\end{tabular}
\caption{Ablation results on HELMET, averaged over context lengths from 8K to 128K. LongQA and Summarization are excluded due to the cost of LLM-as-a-judge evaluation over long outputs.}
\label{tab:ablations-helmet}
\end{table}

We did not evaluate the ablation models on HELMET's LongQA and Summarization categories due to the cost of LLM-as-a-judge evaluation over long outputs.  Tables~\ref{tab:ablations-ruler} and~\ref{tab:ablations-helmet} report results on RULER and the remaining HELMET categories, respectively.  The patterns are broadly consistent with the in-domain findings in Section~\ref{sec:ablations}.  On RULER, `No Recall Reward' drops 4.9 points on average while No Mask loses only 0.8, confirming that gold document supervision is the more important training signal for retrieval-heavy synthetic tasks.  On HELMET, `No Recall Reward' again underperforms, most notably on Re-rank (25.3 vs.\ 46.2), while No Mask matches or exceeds the full model on RAG, ICL, and Cite, but drops substantially on Recall (87.7 vs.\ 96.2), reinforcing the role of constrained decoding for faithful retrieval.  Interestingly, constrained decoding drops slightly on HELMET Cite (13.7 vs.\ 15.4 for No Mask) despite helping on the in-domain entity citation task (61.8 vs.\ 58.1).  This may partly reflect differences in evaluation: the in-domain task uses citation F1 over exact document identifiers, whereas HELMET's citation evaluation incorporates NLI-based assessment of whether cited passages support the generated claims~\citep{gao-etal-2023-enabling}, which may favor flexible evidence composition.

\section{Extended Related Works}
\label{app:section_7}

\subsection{Long-Context Utilization and Evaluation}
\label{app:long_context_eval}

Effectively utilizing long contexts remains a fundamental challenge. \citet{liu-etal-2024-lost} showed that LLM performance degrades when relevant information is in the middle of the context, and LongPiBench~\citep{tian-etal-2025-distance} extends this to the multi-piece setting, finding that the distance between relevant pieces introduces further biases across 32K to 256K tokens. \projectname{}'s constrained decoding addresses positional bias directly: the logit mask selects valid continuations from anywhere in the searchable context regardless of position, and because recall spans serve as a reasoning aid rather than appearing in the final output, positional bias does not affect the faithfulness of recalled evidence. Long-context evaluation has evolved from Needle-in-a-Haystack~\citep{kamradt2023niah} and its multi-needle extensions~\citep{martin2024multineedle} to comprehensive benchmarks. RULER~\citep{hsieh2024ruler} generalizes NIAH into 13 synthetic tasks spanning retrieval, variable tracking, and aggregation. HELMET~\citep{yen2024helmet} is deliberately broad, drawing tasks from LongBench~\citep{bai2024longbench}, InfiniteBench~\citep{zhang2024infinitebench}, and other sources across seven application-driven categories, finding low cross-category correlation with synthetic benchmarks. We evaluate on both RULER and HELMET because they stress complementary failure modes: RULER tests precise retrieval and aggregation on controlled synthetic tasks, while HELMET tests whether improvements generalize to diverse, challenging real-world applications.

\subsection{Improving Long-Context Capabilities}
\label{app:improving_long_context}

A large body of work targets long-context performance~\citep{lu2025a}. One line of work focuses on context extension, increasing the effective context window of pretrained models. ProLong~\citep{gao2025prolong} continues training Llama-3-8B on a curated long-context data mix at sequence lengths up to 512K tokens, and YaRN~\citep{peng2024yarn} modifies rotary position embeddings to efficiently extend context windows without full retraining. \projectname{} focuses on post-training and is agnostic to the context extension recipe; our post-trained \projectname-Qwen2.5-7B uses YaRN to extend its native context window four-fold.

Another line of work reduces context size through retrieval- or memory-augmented generation. Search-R1~\citep{jin2025searchr1} and R1-Searcher~\citep{song2025r1searcher} train models via RL to autonomously issue search queries during step-by-step reasoning, while WebThinker~\citep{li2025webthinker} and DeepResearcher~\citep{zheng-etal-2025-deepresearcher} extend this paradigm to real web environments. MEM1~\citep{zhou2025mem1} takes a complementary approach, training agents to maintain a compact internal state that is rewritten at each turn, compressing long interaction histories into fixed-size memory to achieve constant memory usage across arbitrarily long interactions. While these methods help manage context size, they are orthogonal to \projectname{} and benefit from LLM agents with stronger long-context performance. Indeed, \citet{lee2024longcontextlanguagemodelssubsume} find that long-context LMs already rival dedicated retrieval pipelines and outperform RAG on cross-document reasoning, underscoring that faithful in-context utilization is the emerging bottleneck.

\subsection{RL for Long-Context Reasoning}
\label{app:rl_long_context}

Among methods that directly train for long-context reasoning, several are closely related to \projectname{}. LoongRL~\citep{wang2026loongrl} synthesizes challenging multi-hop training data via UUID key chains and trains with GRPO, inducing emergent plan-retrieve-reason patterns that generalize from 16K training contexts to 128K evaluation. QwenLong-L1~\citep{wan2025qwenlong} uses progressive context scaling with curriculum-guided RL to adapt short-context reasoning models to long-context settings. ALR$^2$~\citep{li2024alr2} takes a pipeline approach, prompting the model to first retrieve relevant evidence from the context before reasoning over it; this can be viewed as a precursor to \projectname{}'s interleaved retrieval, though the fixed retrieve-then-reason pipeline is brittle for tasks where reasoning must precede or naturally interleave with retrieval.

\projectname{} builds on these methods by adding constrained decoding for faithful in-context retrieval, explicit retrieval quality supervision via $R_{\text{ret}}$, and a training recipe that achieves competitive benchmark scores with shorter contexts and less compute (Section~\ref{sec:related-work}).

\subsection{Constrained Decoding and Copy Mechanisms}
\label{app:constrained_decoding}

Grammar-constrained systems~\citep{willard2023outlines,dong2024xgrammar} and entity-constrained generation~\citep{decao2021genre} use logit masking to enforce structural constraints from a fixed grammar or candidate set. $k$NN-LM~\citep{khandelwal2020knnlm} takes a softer approach, interpolating the LM's next-token distribution with a nearest-neighbor distribution over a precompiled datastore of context representations to bias generation toward memorized contexts. Classical copy mechanisms such as CopyNet~\citep{gu2016copynet} and Pointer-Generator Networks~\citep{see2017pointer} learn a soft copy distribution over source tokens. In contrast to all of these, recall spans are learned, model-initiated actions embedded inside free-form reasoning: the model decides when to invoke them to recover and ground evidence, rather than using constraints to emit structured output or relying on external memory.

\section*{LLM Usage Disclosure}

We used GPT-5.2~\citep{openai2025gpt5} to annotate SFT training data by rewriting teacher-produced reasoning traces with verbatim recall spans (Appendix~\ref{app:sft_annotation}). We also used Gemma-3-27B-IT~\citep{gemma2025gemma3} to identify retrieval attempt points in reasoning traces for the injection analysis in Section~\ref{sec:problem}. No LLMs were used to originate research ideas or to generate evaluation results.

\end{document}